%% file: main.tex
\documentclass[letterpaper]{article} 
\usepackage{aaai2026}  
\usepackage{times}  
\usepackage{helvet}  
\usepackage{courier}  
\usepackage[hyphens]{url}  
\usepackage{graphicx} 
\usepackage{natbib}  
\usepackage{caption} 
\usepackage{algorithm}
\usepackage{algorithmic}

\usepackage{multirow}
\usepackage{amsmath}
\usepackage{cleveref}
\usepackage{amsfonts}
\usepackage{booktabs}
\usepackage{xcolor}

\usepackage{newfloat}
\usepackage{listings}
\DeclareCaptionStyle{ruled}{labelfont=normalfont,labelsep=colon,strut=off} 
\floatstyle{ruled}
\newfloat{listing}{tb}{lst}{}
\floatname{listing}{Listing}
\title{TongUI: Internet-Scale Trajectories from Multimodal Web Tutorials\\ for Generalized GUI Agents}
\author {
    Bofei Zhang\textsuperscript{\rm 2}\equalcontrib,
    Zirui Shang\textsuperscript{\rm 1,\rm 2}\equalcontrib,
    Zhi Gao\textsuperscript{\rm 1,\rm 2,\rm 3}\equalcontrib,
    Wang Zhang\textsuperscript{\rm 2}, Rui Xie\textsuperscript{\rm 2,\rm 4}, Xiaojian Ma\textsuperscript{\rm 2}, Tao Yuan\textsuperscript{\rm 2}, Xinxiao Wu\textsuperscript{\rm 1}, Song-Chun Zhu\textsuperscript{\rm 2,3,5}, Qing Li\textsuperscript{\rm 2}\thanks{Qing Li is the corresponding author.}
}
\affiliations {
    \textsuperscript{\rm 1}Beijing Key Laboratory of Intelligent Information Technology, School of Computer Science \& Technology, \\ Beijing Institute of Technology \\
    \textsuperscript{\rm 2}State Key Laboratory for General Artificial Intelligence, BIGAI \\
    \textsuperscript{\rm 3}School of Intelligence Science and Technology, Peking University\\
    \textsuperscript{\rm 4}Shanghai Jiao Tong University 
    \textsuperscript{\rm 5}Department of Automation, Tsinghua University\\
}

\begin{document}

\maketitle

\begin{abstract}
Building Graphical User Interface (GUI) agents is a promising research direction, which simulates human interaction with computers or mobile phones to perform diverse GUI tasks. However, a major challenge in developing generalized GUI agents is the lack of sufficient trajectory data across various operating systems and applications, mainly due to the high cost of manual annotations. 
In this paper, we propose the TongUI framework that transforms millions of multimodal web tutorials into GUI trajectories for generalized GUI agents. Concretely, we crawl GUI videos and articles from the Internet and process them into GUI agent trajectory data. Based on this, we construct the GUI-Net-1M dataset, which contains 1 million trajectories across five operating systems and over 280 applications. To the best of our knowledge, this is the \textbf{largest open-source GUI trajectory dataset}. 
We develop the TongUI agent by fine-tuning Qwen2.5-VL-3B/7B/32B models on GUI-Net-1M, which shows consistent performance improvements on commonly used grounding and navigation benchmarks, outperforming baseline agents by 10\% on multiple benchmarks, showing the effectiveness of the GUI-Net-1M dataset and underscoring the significance of our TongUI framework. 
\end{abstract}

\begin{links}
    \link{Project}{https://computer-use-agents.github.io/tongui/}
    \link{Code}{https://github.com/TongUI-agent/TongUI-agent}
    \link{Dataset}{https://huggingface.co/datasets/Bofeee5675/}
    {GUI-Net-1M}

\end{links}

\section{Introduction}
\label{sec:intro}

Graphical User Interface (GUI) agents based on large foundation models are designed to automate tasks on digital devices by emulating human interactions with a variety of operating systems and applications~\cite{nguyen2024gui,wang2024gui,zhang2024large}. 
These agents utilize large language models (LLMs) or vision-language models (VLMs) to process visual inputs (screenshots) and textual inputs (accessibility tree and HTML code), and produce corresponding actions, such as clicking buttons, filling forms, and scrolling, to complete GUI tasks~\cite{lee2023explore,hong2024cogagent,you2024ferret}. 
GUI agents significantly enhance human-computer interaction, providing potential applications to various domains, such as software testing, financial services, office assistance, and industrial automation, improving work efficiency and user experience.

Recently, notable efforts have been made for GUI agents by fine-tuning the foundation models~\cite{lin2024showui,qin2025ui,xu2024aguvis}. 
However, collecting sufficient trajectory data for fine-tuning is challenging.
Most existing approaches rely on either manually annotated interaction trajectories~\cite{liu2024visualagentbench} that are high-quality but costly to obtain, or synthetic trajectories generated from large open-source or proprietary LLMs/VLMs, which may lack diversity and accuracy~\cite{qin2023toolllm,liu2024apigen,gao2024multi}. 
Desirable GUI agents need to perform actions across a wide variety of applications, each with unique interaction sequences, while collecting comprehensive and diverse data across different applications and operating systems remains a major challenge. 
Thus, the lack of large-scale, diverse, and well-structured GUI trajectory data continues to be a key bottleneck in developing robust and generalized GUI agents.

In this paper, we propose the TongUI framework that transforms millions of multimodal web tutorials into GUI trajectories for generalized GUI agents,
where `Tong' means generalization in Chinese.
Our key observation is that there are readily available multimodal web tutorials on the Internet about how to control computers and smart mobile phones, offering a wealth of information on interacting with various applications and operating systems.
These tutorials, in either videos or articles with screenshot formats, provide detailed step-by-step instructions on interacting with GUI.
Compared to the aforementioned manually collected or synthetic data pipeline, online tutorials offer easy accessibility, rich information content, and good quality.
Thus, it is a natural idea to convert the diverse multimodal web tutorials into task-solving trajectories for GUI agent tuning.



In doing so, we design the TongUI framework composed of four steps: tutorial crawling, tutorial processing, trajectory generation, and data filtering. 
In the tutorial crawling step, we write some seed tasks and use LLMs to extend them into a wider collection of tasks.
The generated tasks serve as keywords for retrieving content from hosts for online GUI tutorials (such as articles from WikiHow and videos from YouTube).
The tutorial processing step aims to extract textual descriptions and screenshots of multimodal tutorials.
We first obtain textual descriptions of multimodal tutorials via automatic speech recognition (ASR) or captioning, through which task queries and plans are produced using LLMs on the obtained textual descriptions.
Then, we extract salient frames from videos as screenshots of each step, while the images in articles are directly regarded as screenshots.
In the trajectory generation step, we leverage a zero-shot GUI agent to automatically recognize trajectories, including reasoning thoughts and actions between two steps.
In the data filtering step, we apply a multi-stage pipeline to ensure data quality and GUI relevance, including duplicate tutorial removal, LLM-based content filtering, and trajectory-level filtering using GUI agents.
Based on the TongUI framework, we construct a GUI-Net-1M dataset that contains 1M trajectories across five operating systems with more than 280 applications. As far as we know, GUI-Net-1M is the biggest open-source GUI trajectory dataset.
Notably, our dataset contains tutorials from different time periods, where the same application may have evolving layouts and task-solving solutions, 
helping the model generalize to diverse GUI environments.

Based on GUI-Net-1M, we develop the TongUI agent using Qwen2.5-VL-3B/7B models.
We evaluate the TongUI agents on both the offline and online settings, and the results show that the TongUI agent exhibits consistent improvements in grounding and navigation capabilities.
The results demonstrate 
underscores the effectiveness of the TongUI framework and the collected GUI-Net-1M dataset that improve the agents in a low-cost manner without the need for expensive manual annotation.



In summary, our contributions are three-fold:
(1) We construct GUI-Net-1M, having 1M trajectories across five operating systems and over 280 applications. As far as we know, it is the biggest open-source GUI trajectory dataset.
(2) We propose the TongUI framework that enables GUI agents to automatically learn from rich web resources, leading to better generalization. 
(3) We develop the TongUI agent by fine-tuning Qwen2.5-VL models on GUI-Net-1M, achieving improvements on multiple popular benchmarks. 


\section{Related Work}

\subsection{GUI Agent}

The advancements of LLMs and VLMs accelerate the development of GUI agents. In the early state, only
closed-source models are used (\emph{e.g.}, GPT-4, claude 3.5, and ChatGLM). Recently, more and more open-source models are released, such as Show-UI~\cite{lin2024showui}, UI-TARS~\cite{qin2025ui}, and CogVLM~\cite{wang2023cogvlm}. 
Grounding and planning are two important capabilities in GUI agents. 
The grounding capability means whether the GUI could identify correct buttons or regions for operations.
Some methods improve the grounding capability by adding more prompts (accessibility trees, HTML, and set-of-mask)~\cite{zhou2023webarena,deng2023mind2web,gur2023real,zhang2024ufo,lu2024omniparser} and collecting grounding data for fine-tuning~\cite{yang2024aria}.
As for the planning capability of GUI agents, existing efforts improve it by prompt engineering~\cite{jia2024agentstore,agashe2024agent}, supervised fine-tuning~\cite{lin2024showui,xu2024aguvis,hong2024cogagent,qin2025ui}, or reinforcement learning~\cite{lai2024autowebglm,putta2024agent}.
In addition, some efforts focus on evaluating the GUI agents from multiple aspects, including action sequence~\cite{rawles2024androidinthewild,he2024webvoyager}, grounding precision~\cite{cheng2024seeclick,liu2024visualwebbench}, trajectory effectiveness~\cite{lin2024videogui,li2025effects}, and task completion~\cite{rawles2024androidworld,trivedi2024appworld,xie2025osworld}. The evaluation performance in turn guides the research of intelligent agents.

\subsection{Agent Data Collection}

Compared with data collection for LLMs or VLMs, collecting high-quality data for agents is more challenging, since agents usually require long trajectories that solve complex tasks in different domains (\emph{e.g.}, GUI, multimodal reasoning, embodied AI).
Commonly used schemes include human annotation~\cite{liu2024visualagentbench,wang2025opencuaopenfoundationscomputeruse} and model synthesis~\cite{qin2023toolllm,liu2024apigen,gao2024multi}.
including the planning, action sequence, and action position of GUI agents~\cite{deng2023mind2web,rawles2024androidinthewild,chen2024guicourse,qin2025ui,cheng2024seeclick,lu2024gui}.
Note that the recent state-of-the-art method UI-TARS~\cite{qin2025ui} has not released its data. In contrast, we have fully released our data.

In addition to the two schemes, considering that there are abundant resources for GUI operations on the Internet, some efforts focus on collecting trajectory data of GUI agents from the Web, to preserve the data quality and reduce costs simultaneously~\cite{ou2024synatra,xu2024agenttrek}. They collect textual tutorials from the Internet, and convert them into trajectories by using LLMs or exploring in simulators. Unlike them, we directly collect multimodal tutorials from the Internet and propose a multimodal tutorial process pipeline to convert them into trajectories without any simulators. 
Meanwhile, compared with synthetic data in simulators, our data is closer to tasks in the real world with practical purposes, benefiting to the generalization capability of GUI agents. In~\cref{label:dataset_comparison}, we show the comparisons of GUI-Net-1M with existing GUI trajectory datasets, including GUI Odessey~\cite{lu2024gui}, AgentTrek~\cite{xu2024agenttrek}, AndroidControl~\cite{li2025effects}, AGUVIS~\cite{xu2024aguvis}, LBI~\cite{su2025learn}, E-ANT~\cite{wang2024ant}, ShowUI~\cite{lin2024showui}, and AITW~\cite{rawles2023androidinthewild}, where GUI-Net-1M has obvious advantages in data size, platform, and operating system numbers.

\begin{table}[ht]
\centering
\small
\begin{tabular}{c|c|c|c}
\toprule
\textbf{Datasets} & \textbf{Size} & \textbf{Platform} & \textbf{OS}\\
\midrule
GUI Odyssey & 7K & W+M & A \\
AgentTrek & 10.4K & W & A+W+L+M+I \\
AndroidControl & 15.3K & W+M & A \\
AGUVIS & 35K & W+M & A+W+L \\
LBI & 42.6K & D+W & L \\
E-ANT & 49K  & M & A \\
ShowUI & 137K & W+M & W+L+I \\
AITW & 715K & W+M & A\\
\textbf{GUI-Net-1M} & 1M & D+W+M & A+W+L+M+I \\
\bottomrule
\end{tabular}
\caption{Comparison of GUI-Net-1M with other GUI trajectory datasets. For platform, `D', `W', and `M' mean the desktop, web, and mobile, respectively. For operating systems (OS), `A, `W', `L', `M', and `I' denote Android, Windows, Linux, MacOS, and iOS, respectively.}
\label{label:dataset_comparison}
\end{table}



\begin{figure*}
    \centering
    \includegraphics[width=0.95\linewidth]{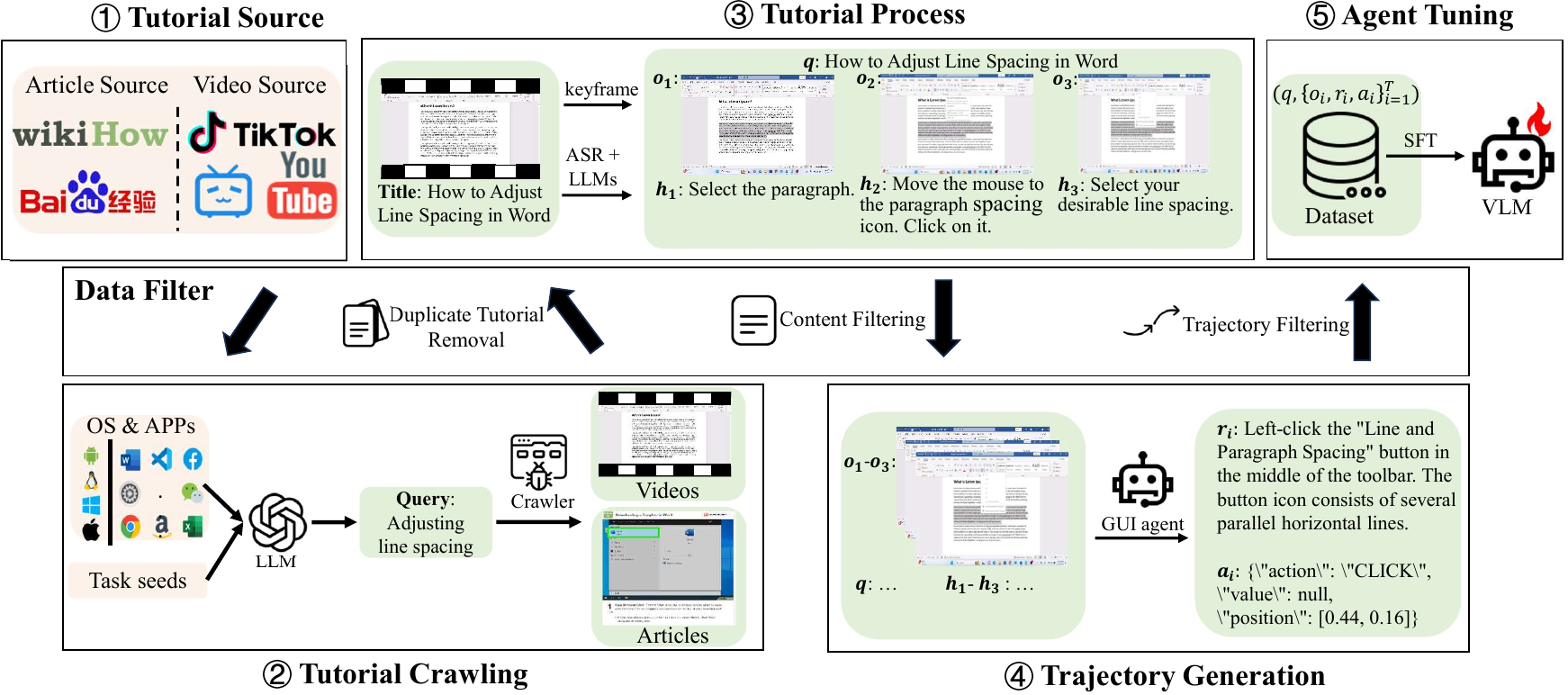}
    \caption{Illustration of the TongUI framework.
    }
    \label{fig:illustration}
\end{figure*}

\section{Method}

\subsection{Formulation}

We formulate GUI tasks as a sequential decision process, and we adopt ReAct~\cite{yao2023react} as our agent framework. At one time step \( i \), given the observation of previous $n$ steps \( o_{i-n}, \cdots, o_{i-1}, o_i \) (\emph{e.g.}, a screenshot of GUI), the thoughts $r$ and actions $a$ of previous $n$ steps $(r_{i-n}, a_{i-n}, \cdots, r_{i-1}, a_{i-1})$, and the query $q$, the agent generates a new thought \( r_i \) and an executable action \( a_i \) from the action space, such as clicking on a specific UI element, entering text, or scrolling through the interface. 
Then, executing the action \( a_i \) leads to a new observation \( o_{i+1} \) (\emph{e.g.}, an updated screenshot). 
This interaction loop continues, with the agent repeatedly observing the environment, selecting actions, and receiving updated observations until either a termination condition is met (\emph{e.g.}, task completes or fails) or the maximum number of steps is reached. 
We parameterize the GUI agent using a vision-language model $M_{\theta}$, 
\begin{align}
r_i^{\star}, a_i^{\star} &= \arg \max_{r_i, a_i} 
    M_{\theta}(r_i, a_i \mid q, 
    o_{i-n}, r_{i-n}, a_{i-n}, \nonumber \\
    &\quad\quad\quad\quad\quad\quad \ldots, 
    o_{i-1}, r_{i-1}, a_{i-1}, o_i).
\end{align}


We propose the TongUI framework that aims to tune $M_{\theta}$ by learning from multimodal web tutorials, as shown in~\cref{fig:illustration}. Such tutorials are usually formatted as $\{v, e\}$, where $v$ denotes the visual information (such as images or videos), and $e$ denotes the textual information (such as an introduction to a video). Our goal is to collect multimodal tutorials $\{v, e\}$, and convert them into training data $\{q, o_1, r_1, a_1, \cdots, o_T, r_T, a_T \}$ with the query $q$ and the trajectory $(o_1, r_1, a_1, \cdots, o_T, r_T, a_T)$ in $T$ steps.

\subsection{Tutorial Source}
To collect multimodal web tutorials, we carefully select multiple data sources via brainstorming, ensuring that we cover a wide range of applications, operating systems, and task types.
Concretely, we choose YouTube, Bilibili, Wikihow, and Baidu Experience, which host much user-generated content on GUI tasks.
YouTube and Bilibili have a variety of videos about GUI tasks.
For video tutorials, the visual information $v$ is the video itself, and the textual information $e$ includes the title, the brief introduction to the video, and the caption or audio of the video.
Wikihow and Baidu Experience are two widely used platforms that provide step-by-step articles with images in each step across applications and operating systems. Here, the visual information $v$ is the image sequence among all steps, and the textual information $e$ includes the title and textual content in the article.
In this case, we could get diverse multimodal web tutorials.


\subsection{Tutorial Crawling}
We use a keyword-based search approach. 
The search keywords are constructed by combining the name of the applications or website with the task content, that is ``{app/web} + task''.
Here, ``app/web'' refers to the operating system or application being used (\emph{e.g.}, a mobile app or website), while ``task'' has a specific objective (\emph{e.g.}, changing font size in Word, or browsing in Chrome). 
We identify task seeds across multiple applications via brainstorming, 
and expanded them as search keywords using LLMs for more diverse and relevant tasks. 

With these keywords, we crawl multimodal web tutorials from source websites.
We employ platform-specific methods to retrieve tutorials, including subtitles and audio. 
For YouTube, we utilize the Google API and YouTube's official API to search for videos, and download the relevant content. 
For Bilibili and TikTok, we use the unofficial API to search for videos and retrieve both video and audio streams for download. For Wikihow and Baidu Experience, articles usually have multiple tags to indicate the main category of the tutorial, such as Windows, Chrome, Word, and \emph{etc.} We crawl text-image tutorials using tags. 

\subsection{Tutorial Process}

\textbf{Textual processing.} 
Given a crawled multimodal tutorial $\{v,e\}$, this process extracts the task $q$ and rough descriptions $\{ h_1, \cdots, h_T\}$ in $T$ steps.
For video tutorials from Youtube, Tiktok, and Bilibili, the task guidance is usually in the video's audio information without any textual introductions.
To solve this issue, we begin by applying the open-source speech recognition model, Whisper~\cite{radford2022robustsr}, to transcribe the audio streams into the text information $e$. 
For image-text articles on Wikihow and Baidu Experience, we parse websites based on their article structures for the textual information $e$.

We use LLMs 
to identify key task-related verbs and nouns, such as ``clicking a button'', ``filling out a form'', or ``selecting a menu''.
Then, we leverage both the title and extracted text to classify tutorials into three categories: mobile, desktop, and others. The data with category ``others'' is discarded.
Finally, we use LLMs to extract the task $q$ and summarize this process into rough descriptions $\{ h_1, \cdots, h_T\}$ of $T$ steps from the filtered textual information.

\textbf{Visual processing.} 
This process extracts the observation $\{ o_1, \cdots, o_T\}$ in the $T$ steps. 
For text-image articles, we could directly obtain the image sequences as the observation $\{ o_1, \cdots, o_T\}$ from the crawled data.
In practice, there are some noisy data in Baidu Experience and Wikihow, which might be diagrams, comics, or natural images rather than GUI screenshots. 
To address this issue, we prompt GPT-4o-mini to classify each image as a screenshot or not.

For video tutorials, we need to extract key frames that correspond to critical actions, representing meaningful moments in the task-solving process.
We observe that the audio transcript of the videos often contains valuable planning information, such as the description of task steps. Therefore, when audio is available, we first segment the video based on the audio transcript timestamps, treating each segment as a potential task step. We then extract key frames from each of these segments.
In the absence of audio, we treat the entire video as a single segment and extract key frames throughout the video.
In doing so, we adopt the MOG2 algorithm~\cite{zivkovic2004,zivkovic2004recursive,zivkovic2006} to detect significant changes. 

\begin{figure*}
    \centering    \includegraphics[width=0.82\linewidth]{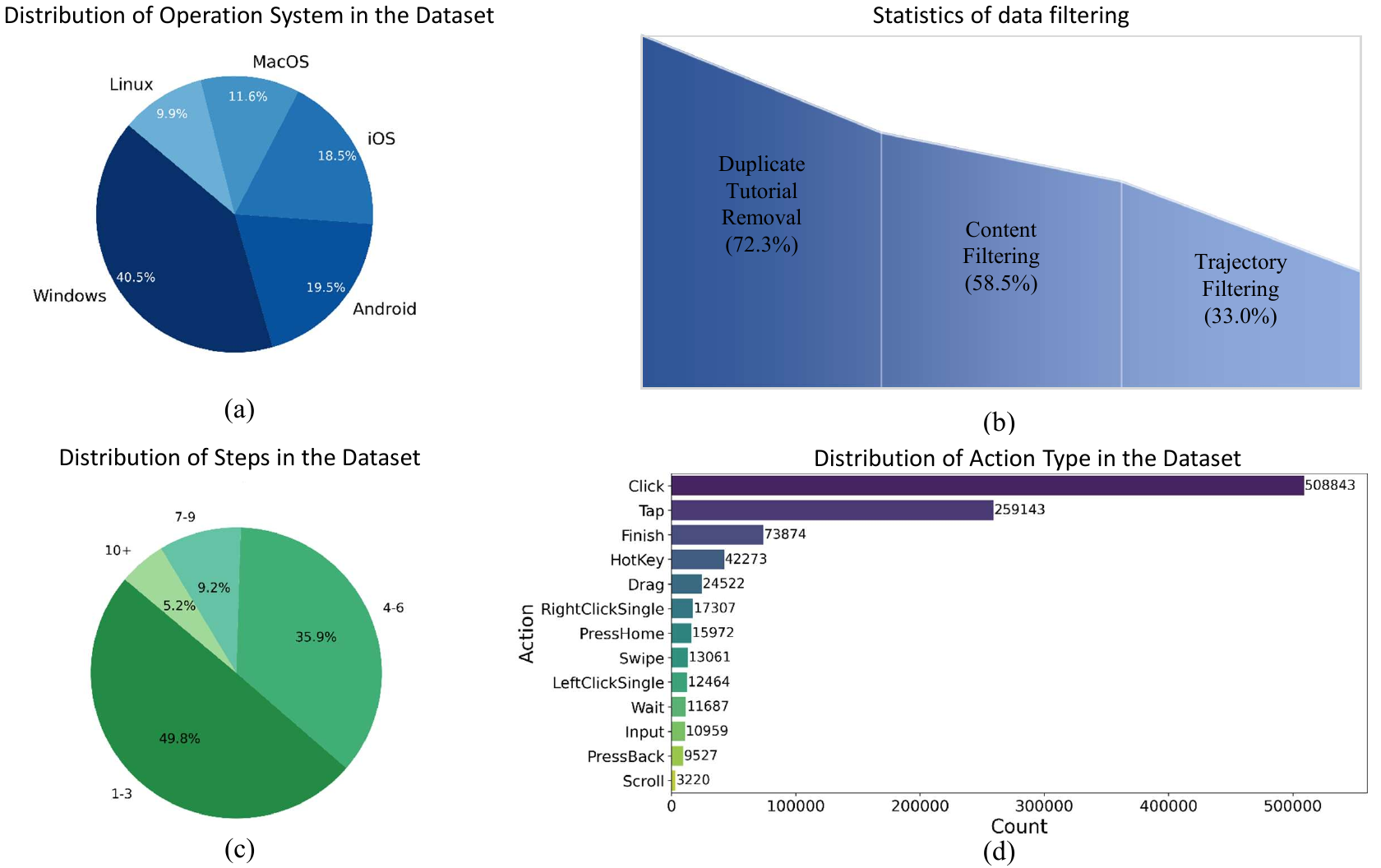}
    \caption{Data statistics on the GUI-Net-1M dataset.}
    \label{fig:statistics}
\end{figure*}

After textual and visual processing, 
tutorials from different sources are structured into a task $q$ with text and image pairs of $T$ steps, denoted as $\big( \{o_1, h_1\},\cdots, \{o_i, h_i\},\cdots, \{o_T, h_T\}\big)$, where $o_i$ is the image and $h_i$ is the rough description in the $i$-th step.

\subsection{Trajectory Generation}
This process aims to generate the trajectory $(o_1, r_1, a_1, \cdots, o_T, r_T, a_T)$ in $T$ steps using the task $q$, and image $o_i$ and rough description $h_i$ in $T$ steps.
Concretely, we use a pretrained GUI agent (such as UI-TARS~\cite{qin2025ui}) to generate the thought $r_i$ and the action $a_i$ using the same action space defined by the GUI agent.
For the thought $r_i$ and action $a_i$, we feed the observation $o_i$ with $h_i$ as the query to the zero-shot agent. 
Here, we use $h_i$ as the query for agents instead of using the task $q$, because $q$ usually contains an abstract goal instead of specific instructions about what to do on the observation. 
We empirically find that using $h_i$ leads to better performance.
In this case, each step of text and image pairs can be written as $\{o_i, r_i, a_i\}$. 
In practice, the model might fail to generate well-formatted actions. We discard this step and make the next action the beginning of a new trajectory. For example, for a 4-step trajectory which has a failed generation in the second step, we split it into two training trajectories: $(\{o_1, r_1, a_1\})$ and $(\{o_3, r_3, a_3\}, \{o_4, r_4, a_4\})$. Finally, we combine the trajectory with the task $q$ as $(q, \{o_i, r_i, a_i\}_{i=1}^T)$.
We collect these trajectories into a GUI-Net-1M dataset that contains 1M trajectories. 
The data is on five operating systems with more than 280 applications.


\subsection{Data Filtering}
We apply a multi-stage data filtering pipeline for data quality, as shown in~\cref{fig:illustration}.
\textbf{(1) Duplicate Tutorial Removal.}
To eliminate redundancy of crawling, we directly remove exact duplicate tutorials based on their unique identifiers (\emph{e.g.}, video IDs and URLs), ensuring uniqueness.
\textbf{(2) Content Filtering.}
After tutorial processing, we employ an LLM to semantically filter out tutorials that are irrelevant to the GUI tasks. Concretely, based on the article content, titles, and audio transcriptions, the LLM evaluates whether the tutorial is about GUI interactions. 
\textbf{(3) Trajectory Filtering.}
In the trajectory generation stage, since we use UI-TARS~\cite{qin2025ui}, if a step is unrelated to GUI interaction, the agent typically predicts the action \texttt{wait} or \texttt{call\_user}. We use this as a signal to discard such observations.
Then, we feed the screenshot and trajectory of a task into a Qwen2.5-VL-7B model, and prompt it to filter out low-quality data.

\subsection{Agent Tuning}

Given a data point  
$(q, \{o_i, r_i, a_i\}_{i=1}^T)$ of $T$-step, we train a VLM $M_{\theta}$ via supervised fine-tuning (SFT),
\begin{align}
\min\; \mathbb{E}_{(q, \{o_i, r_i, a_i\}_{i=1}^T) \sim \mathbb{D}} 
\Big[ - \sum_{i=1}^{T} M_{\theta}(r_i, a_i \mid q, 
o_{i-n}, r_{i-n},  \nonumber  \\ 
a_{i-n}, \ldots, o_{i-1}, r_{i-1}, a_{i-1}, o_i) \Big]
\end{align}
where $\mathbb{D}$ is the collected GUI-Net-1M dataset, and we sum the loss values of the $T$ steps in the trajectory. 
After training, we obtain the TongUI agent.

\section{Dataset Statistics}

We provide four key statistics
to show the diversity of the collected GUI-Net-1M dataset.
We show the \textbf{operating system distribution} in ~\cref{fig:statistics}(a). Our dataset covers a diverse range of operating systems, including Windows, Android, iOS, MacOS, and Linux, ensuring a broad representation of GUI interactions across both desktop and mobile environments.
We provide statistics on the data filtering flow in ~\cref{fig:statistics}(b). Throughout the data filtering pipeline, we observe a gradual reduction in data volume, reflecting the progressive refinement of the dataset. 
After applying duplicate tutorial removal, content filtering, and trajectory filtering, 33.0\% of the original data is retained. This three-step filtering process helps to include high-quality and GUI-relevant interactions in the final dataset.
We show the \textbf{step distribution} in ~\cref{fig:statistics}(c). Our dataset includes GUI interactions with varying step lengths, ranging from single-step actions to 9-step tasks. The distribution shows that shorter tasks (1-3 steps) are more frequent, while longer tasks gradually decrease in numbers. 
The higher proportion of trajectories of 1-3 steps is partly due to the trajectory splitting strategy described in Section 3.5.
The \textbf{action distribution} is shown in ~\cref{fig:statistics}(d).
Click and Tap are the most frequent, reflecting their central role in both desktop and mobile interactions. 

\begin{table*}[h!]
\centering
\small
\begin{tabular}{l|ccc|ccc|ccc}
\hline
\textbf{Agent Model} & \multicolumn{3}{c|}{\textbf{ScreenSpot Avg}} & \multicolumn{3}{c|}{\textbf{ScreenSpot-V2 Avg}} & \multicolumn{3}{c}{\textbf{ScreenSpot-Pro Avg}} \\ 
 & \textbf{Text} & \textbf{Icon} & \textbf{Avg} & \textbf{Text} & \textbf{Icon} & \textbf{Avg} & \textbf{Text} & \textbf{Icon} & \textbf{Avg} \\ \hline
GPT-4o  &17.3  & 18.8 & 18.3 &-&-&-& 1.3  & 0.0  & 0.8  \\
SeeClick-9B~\cite{cheng2024seeclick}  & 68.6 & 38.2  & 53.4 & 67.9 & 37.5 & 55.1 & 1.8 & 0.0  & 1.1  \\
OS-Atlas-4B~\cite{wu2024atlas}  & 86.1 & 62.1 & 76.8 & 81.9 & 56.4  & 71.9 & 5.0  & 1.7  & 3.7  \\
OS-Atlas-7B~\cite{wu2024atlas}  & 92.2 & 75.1 & 85.1 & 92.2 & 72.2 & 84.1 & {28.1} & 4.0  & {18.9} \\
ShowUI-2B~\cite{lin2024showui}  & 83.4 &  66.7	 & 75.1 & -& -& - &  10.8 & 2.6  & 7.7  \\
CogAgent-18B~\cite{hong2024cogagent} &-&-&-&- & -& - &  12.0 & 0.8  & 7.7  \\
Aria-UI~\cite{yang2024aria} & 90.7 & 71.4 &82.4&- &- & - &  17.1 & 2.0  & 11.3 \\
AGUVIS-7B~\cite{xu2024aguvis} & 92.6 & 73.3 & 84.4 & - & - & -  & - & - & - \\
UGround-7B~\cite{gou2024navigating}  &79.5 & 68.1  & 70.4 &- &- & - &  25.0 & 2.8  & 16.5  \\
UI-TARS-2B~\cite{qin2025ui} & 89.3 & 73.0	 & 82.3 & 91.0 & 75.3 &  84.7&  39.6 & 8.4 & 27.7 \\
UI-TARS-7B~\cite{qin2025ui}  & \underline{93.5} & \underline{84.8} & \textbf{89.5} & \underline{95.3} & 86.4 & \underline{91.6} &  \underline{47.8} & \underline{16.2} & \underline{35.7} \\
UI-TARS-72B~\cite{qin2025ui}  & 91.1 & \textbf{85.4}  & 88.4 & 92.5 & \textbf{87.3} & 90.3 & \textbf{50.9} &\textbf{17.5} & \textbf{38.1} \\
\hline
Qwen2.5-VL-3B $\dagger$ & 68.4 & 40.8 & 56.5 & 73.6 & 50.1 & 64.2 &  9.1 & 3.3 & 6.9\\
Qwen2.5-VL-7B $\dagger$ &	87.7 & 66.4 & 78.6 & 92.0 & 72.3 & 84.0 &  17.4 & {4.5} & 12.5\\
TongUI-3B& 90.9 & 76.5  & 83.6 & 91.6 & 77.5 &  85.5&  26.4 & 4.1 & 18.0\\
TongUI-7B &91.6 & 80.4 & 86.0 & 93.2 & 83.0 &  88.7&  {35.1} & {8.0} & {24.7}\\
TongUI-32B & \textbf{94.1} & {82.9}  & \underline{88.5} & \textbf{95.8} & \underline{86.7} & \textbf{92.1} &  {45.9} & {12.6} & {33.1}\\
\hline
\end{tabular}%
\caption{Results on ScreenSpot, and ScreenSpot-V2, and ScreenSpot-Pro. $\dagger$ means the results are reproduced. The best method is marked in bold, and the second-best method is underlined.}
\label{table:experiment_ScreenSpot}
\end{table*}

\begin{table*}[t]
\begin{center}
\small
\begin{tabular}{c | c c c c c c}
\hline
\multicolumn{1}{c}{\bf Method} &\multicolumn{1}{c}{\emph{General}} & \multicolumn{1}{c}{\emph{Single}}& \multicolumn{1}{c}{\emph{Web Shopping}}& \multicolumn{1}{c}{\emph{Install}}& \multicolumn{1}{c}{\emph{Google Apps}}& \multicolumn{1}{c}{\emph{Average}}\\ \hline
PaLM2-CoT~\cite{zhang2024you} & – & – & – & – & – & 39.6 \\
OmniParser~\cite{lu2024omniparser} & 48.3 & 57.8 & 51.6 & 77.4 & 52.9 & 57.7 \\
SeeClick-9.6B~\cite{cheng2024seeclick} & 54.0 & 73.7 & 57.6 & 66.4 & 54.9 & 59.3 \\
ShowUI-2B~\cite{lin2024showui} & 63.9 & {77.5} & \underline{66.6} & 72.5 & 69.7 & 70.0  \\
\hline
Qwen2.5-VL-3B$\dagger$ & 20.7 & 31.4 & 17.1 & 16.3 & 16.8 & 20.5 \\
Qwen2.5-VL-7B$\dagger$ & 39.4 & 41.1 & 35.8 & 43.2 & 42.3 & 40.4 \\
Qwen2.5-VL-3B-ShowUI & \underline{66.0} & 74.4 & 65.0  & 74.5 & 70.3 & 70.1 \\
TongUI-3B & 65.6 & 77.0 & 65.8 & \underline{75.1} & \textbf{74.5} & \underline{71.6} \\
TongUI-7B & \textbf{67.6} & \textbf{79.9} & \textbf{69.1} & \textbf{76.3} & \underline{73.5} & \textbf{73.3} \\
TongUI-32B & 64.0 & \underline{78.4} & 65.0 & 74.2 & \underline{73.5} & 71.0 \\
\hline
\end{tabular}
\end{center}
\caption{Results on the AITW. We report results on five splits of AITW and the average scores.}
\label{table:experiment_AITW}
\end{table*}


\section{Experiments}


\subsection{Setting}


\noindent \textbf{Evaluation.} 
We use Qwen2.5-VL-3B/7B/32B as our base VLM model. 
We evaluate the TongUI agent on offline benchmarks: ScreenSpot~\cite{cheng2024seeclick}, ScreenSpot-V2~\cite{wu2024atlas}, ScreenSpot-Pro~\cite{li2025screenspot}, AITW~\cite{rawles2023androidinthewild}, AndroidControl~\cite{li2025effects}, Mind2Web~\cite{deng2023mind2web}, and UI-Vision~\cite{nayak2025ui}. We also evaluate the TongUI agent on an online benchmark: MiniWob~\cite{shi2017world}, following the same setting in SeeClick\cite{cheng2024seeclick}. 

\noindent \textbf{Compared Methods.}
We mainly compare the TongUI agent with open-source methods, including ShowUI~\cite{lin2024showui}, SeeClick~\cite{cheng2024seeclick}, OS-Atlas~\cite{wu2024atlas}, CogAgent~\cite{hong2024cogagent}, AGUVIS~\cite{xu2024aguvis}, AgentTrek~\cite{xu2024agenttrek}, \emph{etc}. We also compare TongUI with UI-TARS~\cite{qin2025ui} that only releases models without training details and training data.
In contrast, we have fully open-sourced our data, code, and models.





\subsection{Main Results}
\subsubsection{Grounding} 
In~\cref{table:experiment_ScreenSpot}, we show the zero-shot grounding performance of TongUI on ScreenSpot, ScreenSpot-V2, and ScreenSpot-pro. Similar to previous works \cite{lin2024showui}, grounding on icons is much harder than grounding on text. 
The collected data leads to significant improvements on the baseline Qwen2.5-VL model. 
Compared with ShowUI, our method has about 5 \% - 20 \% improvements.
TongUI has a comparable performance to UI-TARS, but UI-TARS only releases the models without training details and data.
These results demonstrate that GUI-Net-1M can indeed improve the grounding capability of GUI agents.

\begin{table}[ht]
\centering
\small
\begin{tabular}{c c | c c }
\hline
\textbf{Method} & \textbf{Model} & \emph{High} & \emph{Low} \\
\hline
AITW & PaLM 2L & 19.5 & {56.7} \\
SeeClick & GPT-4-Turbo & 33.9 & 54.3 \\
M3A & GPT-4-Turbo & {42.1} & 55.0 \\
ER & PALM-2S & 19.5 & 45.5 \\
ER & PALM 2L & 33.0 & 45.9 \\
ER & GPT-4 & 32.1 & 51.7 \\
ER & Gemini 1.5 Pro & 24.4 & 50.2 \\
AGUVIS & AGUVIS-7B & 61.5 & 80.5 \\
AGUVIS & AGUVIS-72B & 66.4 & 84.4 \\
UI-TARS & UI-TARS-2B & 68.9 & 89.3\\
UI-TARS & UI-TARS-7B & 72.5 & 90.8 \\
\hline
TongUI & TongUI-3B & \underline{73.3} & \underline{91.5}\\
TongUI & TongUI-7B & \textbf{76.0} & \textbf{91.9} \\
\hline
\end{tabular}
\caption{Step accuracy on AndroidControl.}
\label{table:experiment_AndroidControl}
\end{table}

\begin{table}[t]
\begin{center}
\small
\setlength{\tabcolsep}{1mm}
\begin{tabular}{c|c|c|c}
\hline
\multicolumn{1}{c}{\bf Model} & \multicolumn{1}{c}{\emph Basic} & \multicolumn{1}{c}{\emph{Functional}} & \multicolumn{1}{c}{\emph{Spatial}} \\ 
\hline 
GPT-4o & 1.6 & 1.5 & 1.0 \\
Geimni-Flash-2.0 & 0.5 & 0.4 & 0.1 \\
Claude-3.7-Sonnet & 9.5 & 7.7 & 7.6 \\
ShowUI-2B~\cite{lin2024showui} & 8.1 & 7.7 & 2.1 \\
AriaUI25-3B~\cite{yang2024aria} & 12.2 & 14.0 & 4.0 \\
UGround-7B~\cite{gounavigating} & 15.4 & 17.1 & 6.3 \\
AGUVIS-7B~\cite{xu2024aguvis} & 17.8 & 18.3 & 5.1 \\
CogAgent-9B~\cite{hong2024cogagent} & 12.0 & 12.2 & 2.6\\
UI-TARS-7B~\cite{qin2025ui} & 20.1 & \underline{24.3} & \underline{8.4} \\
\hline
Qwen2.5-VL-7B & 1.2 & 0.8 & 0.5 \\
TongUI-3B & 22.4 & 17.4 & 6.5\\
TongUI-7B & \underline{24.4} & 22.5 & 7.2\\
TongUI-32B & \textbf{24.5} & \textbf{24.8} & \textbf{11.3}\\
\hline
\end{tabular}
\end{center}
\caption{Results on UI-Vision}
\label{table:experiment_UIvision}
\end{table}

\begin{table*}[ht]
\begin{center}
\small
\setlength{\tabcolsep}{1mm}
\begin{tabular}{c | c c c | c c c | c c c }
\hline
\multirow{2}{*}{\bf Method} & \multicolumn{3}{c|}{\bf Cross-Task} & \multicolumn{3}{c|}{\bf Cross-Website } & \multicolumn{3}{c}{\bf Cross-Domain } \\
\cline{2-10}
 & \emph{Elem. Acc} & \emph{OP. F1} & \emph{Step SR} & \emph{Elem. Acc} & \emph{OP. F1} & \emph{Step SR} & \emph{Elem. Acc} & \emph{OP. F1} & \emph{Step SR} \\ 
\hline
CogAgent~\cite{hong2024cogagent} &  22.4 & 53.0 & 17.6 & 18.4 & 42.4 & 13.4 & 20.6 & 42.0 & 15.5 \\
MindAct~\cite{deng2023mind2web} & 55.1 & 75.7 & 52.0 & 42.0 & 65.2 & 38.9 & 42.1 & 66.5 & 39.6 \\
OmniParser~\cite{lu2024omniparser} & 42.4 & 87.6 & 39.4 & 41.0 & 84.8 & 36.5 & 45.5 & 85.7 & 42.0 \\
ShowUI-2B~\cite{lin2024showui} & 39.9 & 88.6 & 37.2 & 41.6 & 83.5 & 35.1 & 39.4 & 86.8 & 35.2 \\
SeeClick-9.6B~\cite{cheng2024seeclick} & 28.3 & 87.0 & 25.5 & 21.4 & 80.6 & 16.4 & 23.2 & 84.8 & 20.8 \\
AgentTrek~\cite{xu2024agenttrek} & 45.5 & 84.9 & 40.9 & 40.8 & 82.8 & 35.1 & 48.6 & 84.1 & 42.1 \\
UI-TARS-2B~\cite{qin2025ui} & 62.3 & 90.0 & 56.3 & 58.5 & 87.2 & 50.8 & 58.8 & 89.6 & 52.3 \\
UI-TARS-7B~\cite{qin2025ui} & \underline{73.1} & \underline{92.2} & \underline{67.1} & \underline{68.2} & \underline{90.9} & \underline{61.7} & \underline{66.6} & \underline{90.9} & \underline{60.5}\\
UI-TARS-72B~\cite{qin2025ui} & \textbf{74.7} & \textbf{92.5} & \textbf{68.6} & \textbf{72.4} & \textbf{91.2} & \textbf{63.5} & \textbf{68.9} & \textbf{91.8} & \textbf{62.1} \\
\hline
Qwen2.5-VL-3B $\dagger$ & 2.5 & 14.5 & 0.4 & 2.7 & 12.6 & 1.0 & 3.3 & 24.2 & 1.7 \\
Qwen2.5-VL-7B $\dagger$ & 6.2 & 72.8 & 5.0 & 6.3 & 68.2 & 4.5 & 8.4 & 73.6 & 7.2 \\
Qwen2.5-VL-3B-ShowUI & 43.2 & 88.7 & 39.7 & 41.3 & 86.7 & 35.5 & 45.1 & 86.1 & 40.7 \\
TongUI-3B & 53.4 & 89.0 & 48.8 & 54.2 & 86.4 & 48.1 & 53.8 & 88.2 & 49.5 \\
TongUI-7B & 58.1 & 88.7 & 53.4 & 55.6 & 87.2 & 49.0 & 57.6 & 88.7 & 52.9 \\
TongUI-32B & 57.2 & 88.1 & 52.4 & 57.4 & 85.8 & 50.6 & 59.2 & 87.8 & 54.1 \\
\hline
\end{tabular}
\end{center}
\caption{Results on Mind2Web. We report results on three types of tasks: cross-task, cross-website, and cross-domain. ``Elem. Acc'' means whether the element is selected correctly, ``OP. F1'' denotes the F1 score for the predicted action, and ``Step SR'' counts successful steps.}
\label{table:experiment_Mind2Web}
\end{table*}

\subsubsection{Offline Navigation}
We evaluate the offline navigation capability of GUI agents on the AITW, AndroidControl, Mind2Web, and UI-Vision datasets, and results are shown in~\cref{table:experiment_AITW},~\cref{table:experiment_AndroidControl},~\cref{table:experiment_Mind2Web}, and~\cref{table:experiment_UIvision}, respectively.
No matter whether on a small model (3B) or a big model (32B), using GUI-Net-1M data leads to competitive performance. 
For example, TongUI-3B achieves better performance compared to ShowUI-2B by 1.6\% on AITW and more than 10\% on UI-Vision.
The larger model, TongUI-7B, gains better performance compared to TongUI-3B, which is consistent with common sense. 
On the AndroidControl and UI-Vision datasets, TongUI achieves even better performance than UI-TARS.
This highlights that the collected data improves the generalization capability of GUI agents. 
We argue that the reason is that the collected data involves diverse applications and operating systems, improving generalization.

\subsubsection{Online Navigation}

We evaluate the online navigation performance on MiniWob, as shown in~\cref{table:experiment_MiniWob}. 
Compared with offline navigation, online navigation is more challenging, since it processes dynamic environments, handles unexpected obstacles, and adapts to changes in the navigation path. In this case, TongUI, which learns from multimodal tutorials, achieves improvements again. 
Considering that the multimodal tutorials are offline data, the improvements confirm the generalization of TongUI.

\begin{table}[t]
\begin{center}
\small
\begin{tabular}{c c | c}
\hline
\multicolumn{1}{c}{\bf Model} & \multicolumn{1}{c}{\bf Finetuned} & \multicolumn{1}{c}{\emph{Score}}\\ \hline 
CC-Net(SL)~\cite{humphreys2022data} & $\checkmark$ &  23.4 \\
Pix2Act~\cite{shaw2023pixels} & $\checkmark$ &  55.2 \\
AGUVIS-72B~\cite{xu2024aguvis} & $\checkmark$ &  66.0 \\
SeeClick-9.6B~\cite{cheng2024seeclick} & $\checkmark$  & 67.0 \\
Qwen2-VL-2B~\cite{wang2024qwen2}& $\checkmark$ & 66.8 \\
ShowUI-2B~\cite{lin2024showui} & $\checkmark$ & 71.5 \\
\hline
Qwen2.5-VL-3B & $\times$ & 0.3 \\
Qwen2.5-VL-3B-ShowUI & $\checkmark$ & 67.7 \\
TongUI-3B & $\checkmark$ & {72.7} \\
TongUI-7B & $\checkmark$ & \underline{73.9} \\
TongUI-32B & $\checkmark$ & \textbf{74.3} \\
\hline
\end{tabular}
\end{center}
\caption{Results on MiniWob}
\label{table:experiment_MiniWob}
\end{table}



\subsection{User Study for Data Quality}

\begin{figure}
    \centering
    \includegraphics[width=0.88\linewidth]{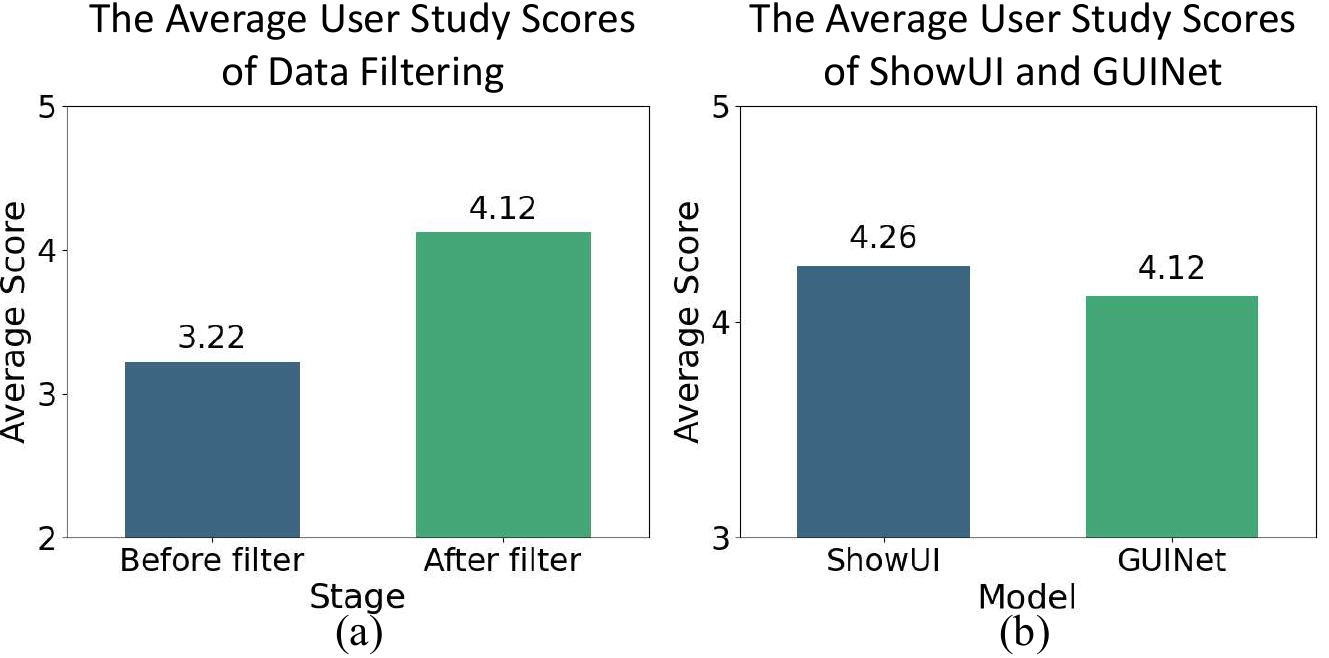}
    \caption{Average scores from humans}
    \label{fig:userstudy}
\end{figure}

To validate the effectiveness of our data filtering strategy and evaluate the data quality of the GUI-Net-1M dataset, we conduct two user studies.
In both studies, five participants with substantial experience in GUI agent research (but not involved in our project) are asked to rate the quality of each data point on a scale from 0 (very poor) to 5 (excellent).
As shown in~\cref{fig:userstudy}(a), we evaluate the impact of our Trajectory Filtering step. Specifically, we randomly sample 100 data points before filtering and 100 after filtering, and mix them. As shown in~\cref{fig:userstudy}(a), the average score increases from 3.22 to 4.12 after filtering, demonstrating that Trajectory Filtering significantly improves data quality by effectively removing low-quality or GUI-irrelevant steps.
We also randomly sample 100 data points from the ShowUI dataset for comparison. The results are shown in~\cref{fig:userstudy}(b), and the ShowUI dataset receives an average score of 4.26.
This suggests that data quality in GUI-Net-1M is comparable to that in the ShowUI dataset, while our data leads to further improvements.

\section{Conclusion}

In this paper, we present the TongUI framework that converts multimodal web tutorials for an 1M trajectory dataset for GUI agents.
By defining suitable and rich sources, we can crawl multimodal tutorials about diverse applications in different operating systems. 
The proposed tutorial processing method can extract tasks and trajectories from tutorials, and we obtain the GUI-Net-1M dataset. 
Using this dataset, the tuned Qwen2.5-VL model achieves improvements on multiple commonly used benchmarks, demonstrating the effectiveness of the TongUI framework and the collected GUI-Net-1M dataset.
In this method, we have to collect all tutorials and train the model once, which may overlook the potential for continual learning. In the future, we will explore the continual learning capability of GUI agents, enabling the agents to better adapt to new environments.

\section{Acknowledgments}
This work was supported by National Science and Technology Major Project (2022ZD0114900), the Natural Science Foundation of China (NSFC) under Grants (No. 62406009), and Opening Project of the State Key Laboratory of General Artificial Intelligence (SKLAGI2024OP01, SKLAGI2024OP14).

\bibliography{main}

\input{supp}

\end{document}

%% file: supp.tex
\newpage
\clearpage

\twocolumn[{%
\renewcommand\twocolumn[1][]{#1}%
\begin{center}
    \centering
    \captionsetup{type=figure}
    \includegraphics[width=1\textwidth]{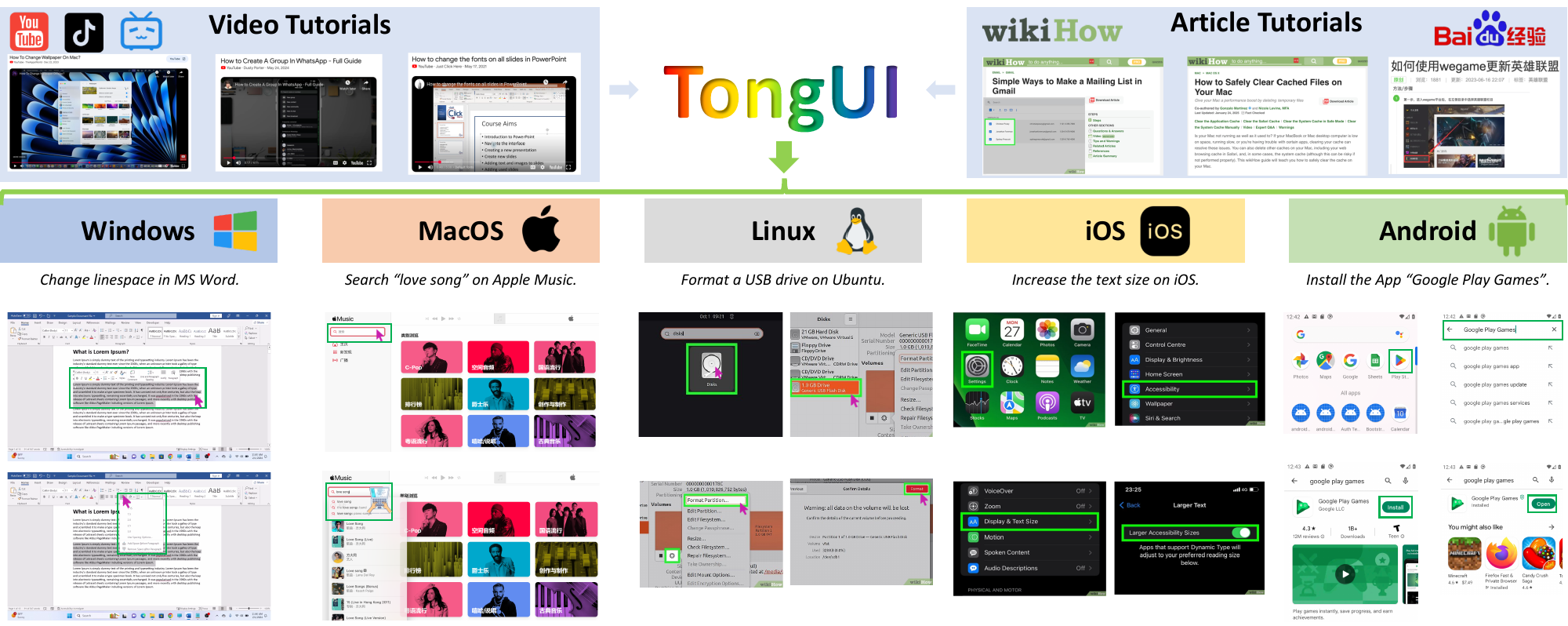}
    \captionof{figure}{
    TongUI crawls and processes multimodal web tutorials from diverse sources into GUI trajectories. The collected GUI-Net-1M dataset spans over 280 applications across five operating systems, improving generalization of GUI agents.  
    }
    \label{fig:illustration_examples}
\end{center}%
}]

\appendix
\newpage
\section{Details}

\subsection{Evaluation Metric.} After training, we perform a zero-shot evaluation of the model's grounding capability on ScreenSpot, ScreenSpot-V2, and ScreenSpot-Pro. The annotation of these grounding benchmarks gives a bounding box for each interactive element. The action from the model is considered correct when the point predicted falls within the bounding box. For online evaluations, the environment of MiniWob provides reward functions, which suggests if a trajectory is correct. For offline evaluations such as Mind2Web, AITW, Baidu Experience, AndroidControl, and GUI Odyssey, the model will be asked to output actions with parameters. We evaluate if the action is correct by the exact match. For parameters, we do the exact match for string parameters and check if a point parameter falls within the bounding box based on annotations. The step is correct if and only if both of the action prediction and parameter prediction are correct.

\subsection{Model Architecture}

\begin{figure}
    \centering
    \includegraphics[width=\linewidth]{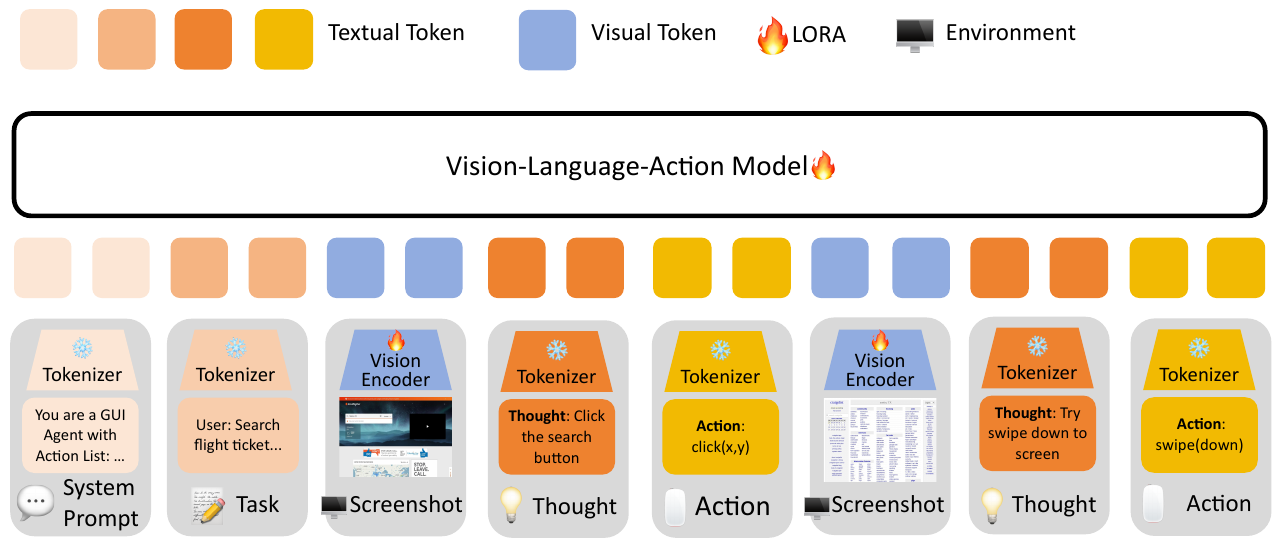}
    \caption{Illustration for TongUI model architecture.}
    \label{fig:network}
\end{figure}

The model architecture is shown in~\cref{fig:network}.

\subsection{Action Space}

The used action space is shown in~\cref{tab:action}.

\begin{table*}
  \centering
  \resizebox{1.8\columnwidth}{!}{
  \begin{tabular}{c c c}
    \hline
    \bfseries Environment & \bfseries Action & \bfseries Definition \\
    \hline
    \multirow{9}{*}{Desktop} & Click(x,y) & Left single click on an element. \\
    & Input(value, x, y) & Type a string into certain element. \\
    & Scroll(direction, x, y) & Scroll screen with left/right/up/down direction. X and y is the start point. \\
    & LeftClickDouble(x, y) & Left double click on an element \\
    & RightClickSingle(x, y) & Right single click on an element \\
    & Drag(x1,y1,x2, y2) & Drag the cursor to the position with left key pressed. \\
    & HotKey(value) & Press the hot key. \\
    & Hover(x, y) & Hover over an element. \\
    & Finish() & Finish the task. \\
    \hline
    \multirow{7}{*}{Mobile} & Tap(x, y) & Finger tap on an element. \\
    & Input(value, x, y) & Type a string into an element. \\
    & Swipe(x1, y1, x2, y2) & Swipe the screen from the start position to the end position. \\
    & LongPress(duration, x, y) & Press on an element with certain seconds. \\
    & PressHome() & Press the home button. \\
    & PressBack() & Press the back button. \\
    & Finish() & Finish the task. \\
    \hline
  \end{tabular}
  }
  \caption{Action space in trajectory generation.}
  \label{tab:action}
\end{table*}

\subsection{Resource Consumption}

\noindent \textbf{Training.} We use Qwen2.5-VL-3B/7B/32B as our base vision-language-action model. 
We train our model with a context window of 8192. We set the maximum number of previous observations as 2, which results in at most three images in a trajectory. We limit each image to 1350 vision tokens. Our training involves two stages. 
Following existing methods~\cite{lin2024showui}, 
we utilize the SFT data (training set) of the evaluation dataset to fine-tune the model. 
We apply LoRA on our model's vision encoder and LLM parts with rank 16 and alpha 32. This setting leads to 0.5\% of total parameters. We set the batch size per device to 4 and the gradient accumulation steps to 8. We set the learning rate as $1e-4$ and AdamW optimizer for training. The training process on 8$\times$A100-80G takes roughly four days for a 3B model and seven days for a 7B model. 

\subsection{Prompt}
In the following, we show the system prompt in our agent framework.
\begin{lstlisting}[caption=System prompt for GUI-Net Agent, label=lst:systemPrompt, numbers=none, breaklines=true, breakindent=0pt]
You are an assistant trained to navigate the computer/mobile screen. 
Given a task instruction, a screen observation, and an action history sequence, output the next action and wait for the next observation. 
Here is the action space:
{action_space}
Format your response as
Thought: <your reasoning process>
Action: <the next action>

Format the action as a JSON object with the following keys:
{"action": "ACTION_TYPE", "value": "element", "position": [x,y]}

You can output multiple actions at once, and use JSON array to represent multiple actions.
If value or position is not applicable, set it as `None`.
Position might be [[x1,y1], [x2,y2]] if the action requires a start and end position.
Position represents the relative coordinates on the screenshot and should be scaled to a range of 0-1.
\end{lstlisting}

In the following, we show a typical action space prompt for desktop.
\begin{lstlisting}[caption=Action space for GUI-Net Agent, label=lst:actionSpace,numbers=none, breaklines=true]
1. `CLICK`: Click on an element, value is not applicable and the position [x,y] is required. 
2. `INPUT`: Type a string into an element, value is a string to type and the position [x,y] is required. 
3. `SCROLL`: Scroll the screen, value is the direction to scroll and the position is start position of the scroll operation.
4. `LEFT_CLICK_DOUBLE`: Left click on an element twice, value is not applicable and the position [x,y] is required.
5. `RIGHT_CLICK_SINGLE`: Right click on an element once, value is not applicable and the position [x,y] is required.
6. `DRAG`: Drag the cursor to the specified position with the left button pressed. Value is not applicable and position [[x1,y1], [x2,y2]] is the start and end position of the drag operation.
7. `HOT_KEY`: Press a hot key, value is the hot key and the position is not applicable.
8. `WAIT`: Wait for 5 seconds, and take a screenshot to check for any changes. Value and position are not applicable.
9. `FINISH`: Finish the task. Value and position are not applicable.
\end{lstlisting}

\section{Dataset}




\subsection{Applications}

\begin{figure*}
    \centering
    \includegraphics[width=0.98\linewidth]{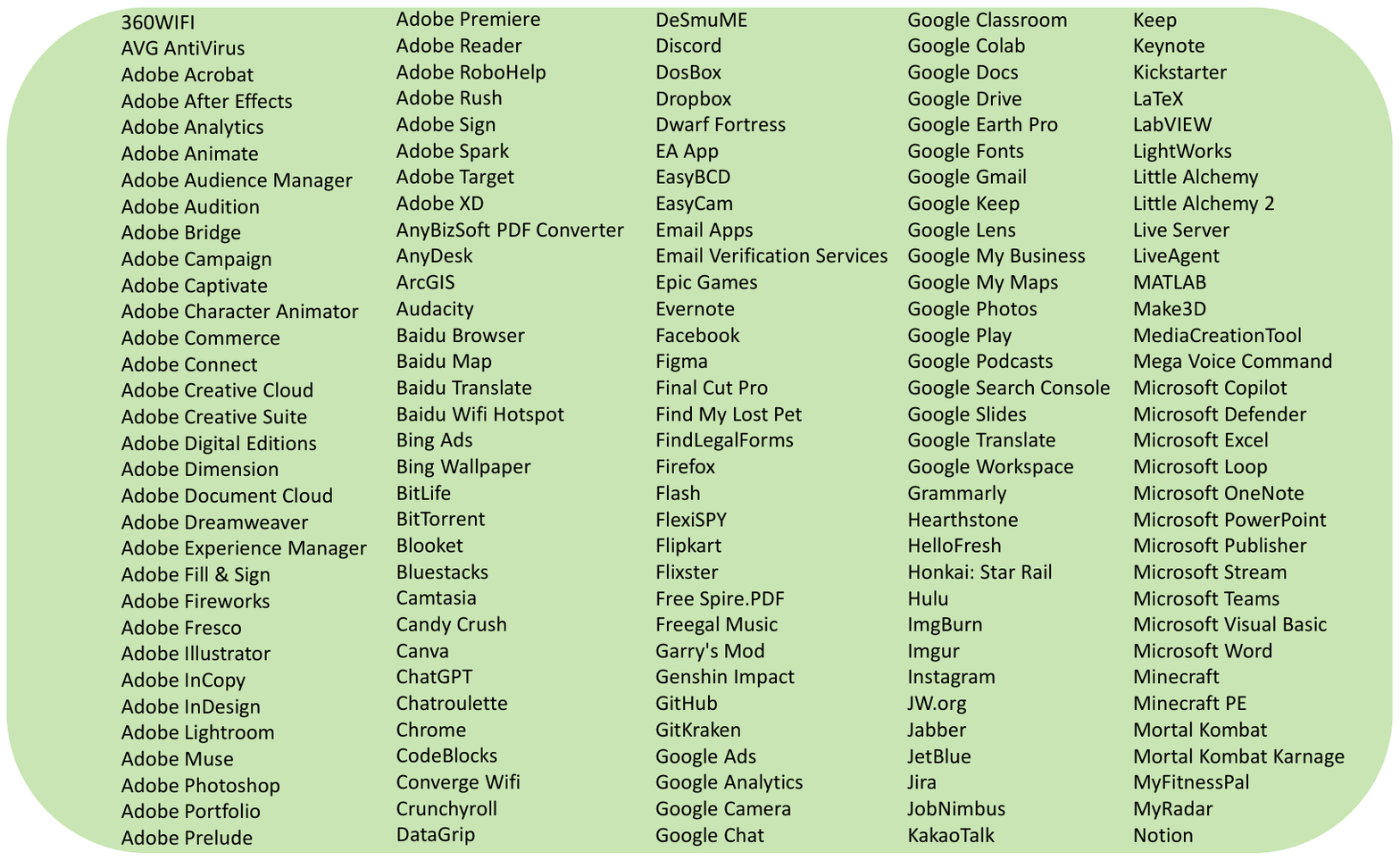}
    \caption{Collected applications.}
    \label{fig:app1}
\end{figure*}

\begin{figure*}
    \centering
    \includegraphics[width=0.98\linewidth]{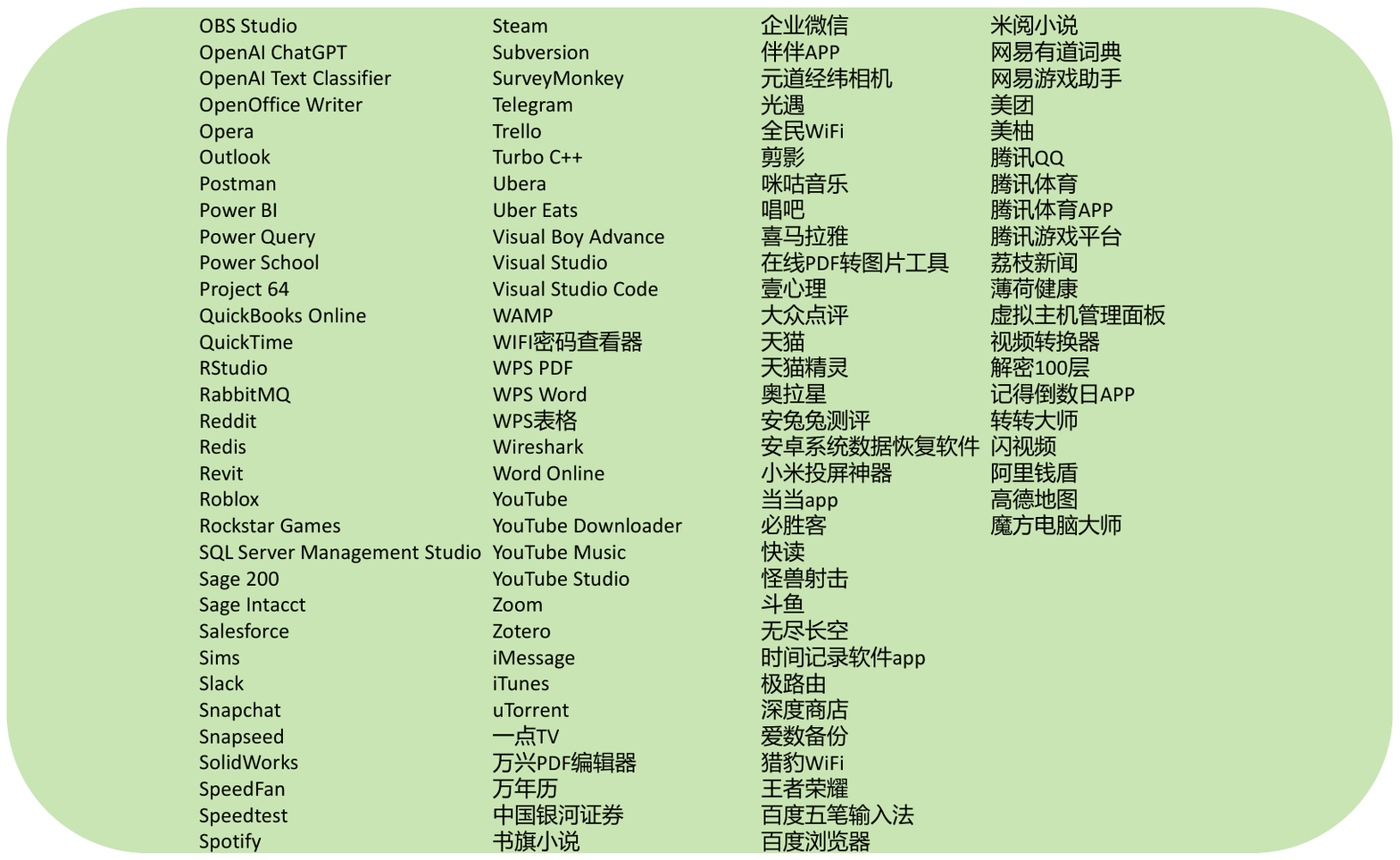}
    \caption{Collected applications.}
    \label{fig:app2}
\end{figure*}

\begin{figure*}
    \centering
    \includegraphics[width=0.98\linewidth]{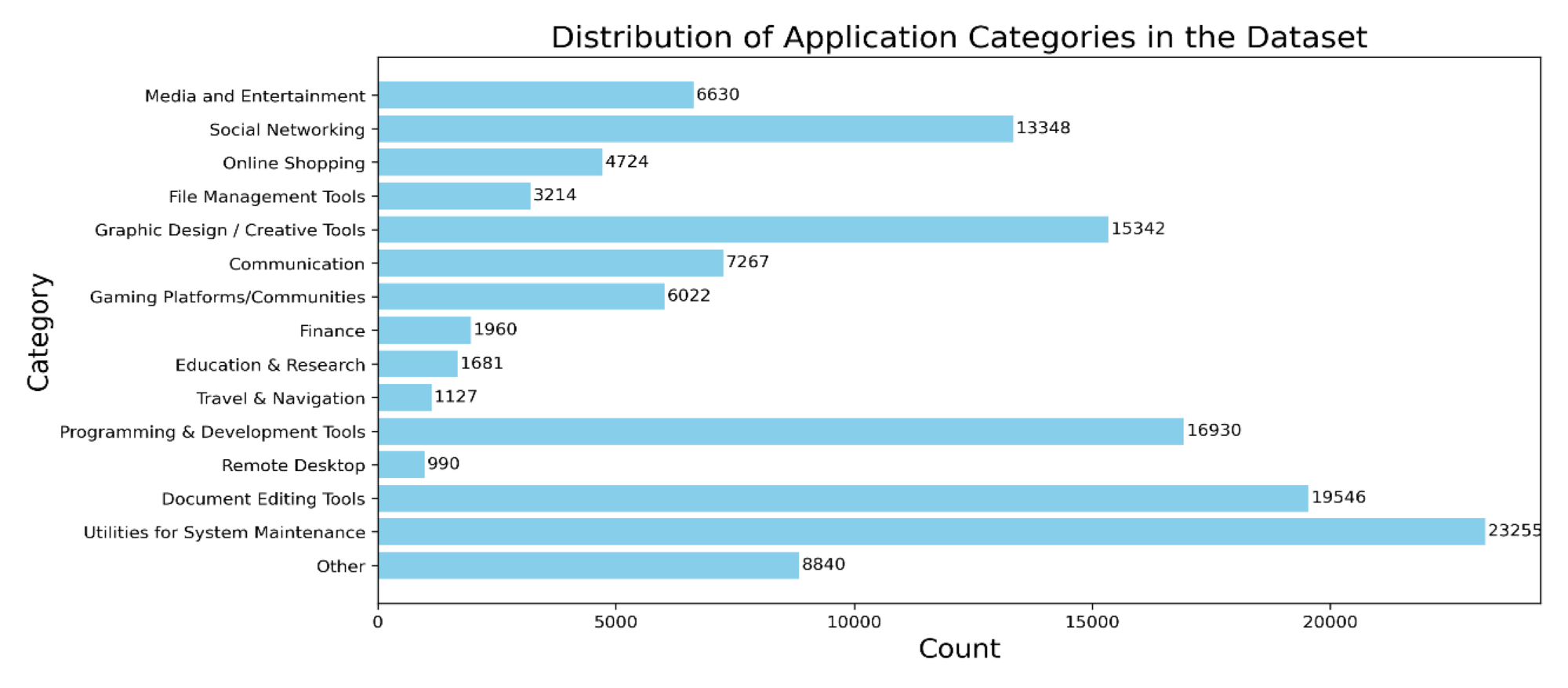}
    \caption{Categories of collected applications.}
    \label{fig:application_statics}
\end{figure*}

In GUI-Net, the collected data spans about 280 applications that are shown in~\cref{fig:app1,fig:app2}.
The distribution of application category is shown in ~\cref{fig:application_statics}. Our dataset covers a wide variety of application categories, ensuring a broad and diverse range of user tasks.

\begin{table*}[t]
\resizebox{0.9\linewidth}{!}{%
\begin{tabular}{c | c c c c c c c}
\hline
\multicolumn{1}{c}{\bf Model}  &  \multicolumn{1}{c}{\emph{Desktop Icon}}& \multicolumn{1}{c}{\emph{Desktop Text}}& \multicolumn{1}{c}{\emph{Mobile Icon}}& \multicolumn{1}{c}{\emph{Mobile Text}}& \multicolumn{1}{c}{\emph{Web Icon}}& \multicolumn{1}{c}{\emph{Web Text}}& \multicolumn{1}{c}{\emph{Average}}\\ \hline
SeeClick-9.6B~\cite{cheng2024seeclick}  & 30.0 & 72.2 & 52.0 & 78.0 & 32.5 & 55.7 & 53.4\\
UGround-7B~\cite{gou2024navigating}  & 63.6 & 82.5 & 60.3 & 82.8 & 80.4 & 73.3 & 70.4 \\
OmniParser-GPT-4V~\cite{lu2024omniparser} & 63.6 & 91.3 & 57.0 & \underline{93.9} & 51.0 & 81.3 & 73.0 \\
ShowUI-2B~\cite{lin2024showui}  & 61.1 & 76.3 & 75.5 & 92.3 & 63.6 & 81.7 & 75.1 \\
AgentTrek~\cite{xu2024agenttrek} & - & - & - & - & 51.5 & 81.7 & -\\
AGUVIS-7B~\cite{xu2024aguvis} & 67.1 & 93.8 & 77.7 & 95.6 & 75.2 & 88.3 & 84.4 \\
UI-TARS-2B~\cite{qin2025ui} & 68.6 & 90.7 & 75.5 & 93.0 & 74.8 & 84.3 & 82.3 \\
UI-TARS-7B~\cite{qin2025ui} & 85.7 & 95.9 & 85.2 & 94.5 & 83.5 & 90.0 & 89.5 \\
UI-TARS-72B~\cite{qin2025ui} & 82.5 & 94.9  & 88.6 & 89.7  & 85.0 & 88.7  & 88.4 \\
\hline
Qwen2.5-VL-3B $\dagger$  & 30.7 & 66.0 & 40.2 & 70.0 & 51.5 & 69.1 & 56.5 \\
Qwen2.5-VL-7B $\dagger$ & 60.0 & 86.6 & 71.6 & 91.2 & 67.5 & 85.2 & 78.6 \\ 
TongUI-3B  & 77.1 & 92.3 & 77.7 & 92.6 & 74.8 & 87.8 & 83.6 \\
TongUI-7B  & \underline{80.0} & \underline{93.8} & \underline{79.5} & 91.9 & \underline{81.6} & \underline{89.1} & \underline{86.0} \\
TongUI-32B  & \textbf{80.0} & \textbf{94.8} & \textbf{84.3} & \textbf{96.3} & \textbf{84.5} & \textbf{91.3} & \textbf{88.5} \\
\hline
\end{tabular}
}
\vskip -0.1in
\caption{Results on ScreenSpot; $\dagger$ means the results are re-produced. 
The best method is marked in bold, and the second-best method is underlined.}
\label{table:supp_experiment_ScreenSpot}
\vskip -0.1in
\end{table*}

\begin{table*}[htbp]
\centering
\small
\resizebox{0.75\textwidth}{!}{%
\begin{tabular}{lccccccccc}
\hline
\textbf{Model} & \multicolumn{2}{c}{\textbf{Mobile}} & \multicolumn{2}{c}{\textbf{Desktop}} & \multicolumn{2}{c}{\textbf{Web}} & \textbf{Avg} \\
& \textbf{Text} & \textbf{Icon/Widget} & \textbf{Text} & \textbf{Icon/Widget} & \textbf{Text} & \textbf{Icon/Widget} & \\
\hline
SeeClick~\cite{cheng2024seeclick} & 78.4 & 50.7 & 70.1 & 29.3 & 55.2 & 32.5 & 55.1 \\
OS-Atlas-4B~\cite{wu2024atlas} & 87.2 & 59.7 & 72.7 & 46.4 & 85.9 & 63.1 & 71.9 \\
OS-Atlas-7B~\cite{wu2024atlas} & 95.2 & 75.8 & 90.7 & 63.6 & \underline{90.6} & 77.3 & 84.1 \\
UI-TARS-2B~\cite{qin2025ui} & 95.2 & 79.1 & 90.7 & 68.6 & 87.2 & 78.3 & 84.7 \\
UI-TARS-7B~\cite{qin2025ui} & 96.9 & 89.1 & 95.4 & 85.0 & 93.6 & 85.2 & 91.6 \\
UI-TARS-72B~\cite{qin2025ui} & 94.8 & 86.3 & 91.2 & 87.9 & 91.5 & 87.7 & 90.3 \\
\hline
Qwen2.5-VL-3B $\dagger$ & 77.2 & 54.5 & 74.7 & 35.7 & 68.8 & 60.1 & 64.2 \\
Qwen2.5-VL-7B $\dagger$ & \underline{95.5} & 81.5 & 90.2 & 65.0 & 90.2 & 70.4 & 84.0 \\
TongUI-3B & 94.4 & 79.6 & 92.8 & 75.0 & 87.6 & 77.8 & 85.5 \\
TongUI-7B & 93.1 & \underline{81.5} & \underline{96.4} & \underline{82.9} & 90.2 & \underline{84.7} & \underline{88.7} \\
TongUI-32B & \textbf{97.9} & \textbf{86.7} & \textbf{96.9} & \textbf{86.7} & \textbf{92.7} & \textbf{86.7} & \textbf{92.1} \\
\hline
\end{tabular}
}
\caption{Results on ScreenSpot-V2.}
\label{table:supp_experiment_ScreenSpotV2}
\end{table*}


\begin{table*}[h!]
\centering
\resizebox{\textwidth}{!}{%
\begin{tabular}{l|ccc|ccc|ccc|ccc|ccc|ccc|ccc}
\hline
\textbf{Agent Model} & \multicolumn{3}{c|}{\textbf{Development}} & \multicolumn{3}{c|}{\textbf{Creative}} & \multicolumn{3}{c|}{\textbf{CAD}} & \multicolumn{3}{c|}{\textbf{Scientific}} & \multicolumn{3}{c|}{\textbf{Office}} & \multicolumn{3}{c|}{\textbf{OS}} & \multicolumn{3}{c}{\textbf{Avg}} \\ 
 & \textbf{Text} & \textbf{Icon} & \textbf{Avg} & \textbf{Text} & \textbf{Icon} & \textbf{Avg} & \textbf{Text} & \textbf{Icon} & \textbf{Avg} & \textbf{Text} & \textbf{Icon} & \textbf{Avg} & \textbf{Text} & \textbf{Icon} & \textbf{Avg} & \textbf{Text} & \textbf{Icon} & \textbf{Avg} & \textbf{Text} & \textbf{Icon} & \textbf{Avg} \\ \hline
GPT-4o & 1.3  & 0.0  & 0.7  & 1.0  & 0.0  & 0.6  & 2.0  & 0.0  & 1.5  & 2.1  & 0.0  & 1.2  & 1.1  & 0.0  & 0.9  & 0.0  & 0.0  & 0.0  & 1.3  & 0.0  & 0.8  \\
SeeClick~\cite{cheng2024seeclick}         & 0.6  & 0.0  & 0.3  & 1.0  & 0.0  & 0.6  & 2.5  & 0.0  & 1.9  & 3.5  & 0.0  & 2.0  & 1.1  & 0.0  & 0.9  & 2.8  & 0.0  & 1.5  & 1.8  & 0.0  & 1.1  \\
OS-Atlas-4B~\cite{wu2024atlas}  & 7.1  & 0.0  & 3.7  & 3.0  & 1.4  & 2.3  & 2.0  & 0.0  & 1.5  & 9.0  & 5.5  & 7.5  & 5.1  & 3.8  & 4.8  & 5.6  & 0.0  & 3.1  & 5.0  & 1.7  & 3.7  \\
ShowUI-2B~\cite{lin2024showui}  & 16.9 & 1.4  & 9.4  & 9.1  & 0.0  & 5.3  & 2.5  & 0.0  & 1.9  & 13.2 & 7.3  & 10.6 & 15.3 & 7.5  & 13.5 & 10.3 & 2.2  & 6.6  & 10.8 & 2.6  & 7.7  \\
CogAgent-18B~\cite{hong2024cogagent} & 14.9 & 0.7  & 8.0  & 9.6  & 0.0  & 5.6  & 7.1  & 3.1  & 6.1  & 22.2 & 1.8  & 13.4 & 13.0 & 0.0  & 10.0 & 5.6  & 0.0  & 3.1  & 12.0 & 0.8  & 7.7  \\
Aria-UI~\cite{yang2024aria}  & 16.2 & 0.0  & 8.4  & 23.7 & 2.1  & 14.7 & 7.6  & 1.6  & 6.1  & 27.1 & 6.4  & 18.1 & 20.3 & 1.9  & 16.1 & 4.7  & 0.0  & 2.6  & 17.1 & 2.0  & 11.3 \\
UGround-7B~\cite{gou2024navigating}  & 26.6 & \underline{2.1}  & 14.7 & 27.3 & 2.8 & 17.0 & \underline{14.2} & 1.6  & \underline{11.1} & 31.9 & 2.7  & 19.3 & 31.6 & \underline{11.3} & 27.0 & 17.8 & 0.0  & 9.7  & 25.0 & 2.8  & 16.5  \\
OS-Atlas-7B~\cite{wu2024atlas}      & \underline{33.1} & 1.4  & \underline{17.7} & \underline{28.8} & 2.8 & \underline{17.9} & 12.2 & \underline{4.7}  & 10.3 & 37.5 & 7.3  & 24.4 & \underline{33.9} & 5.7  & \underline{27.4} & \underline{27.1} & \underline{4.5}  & \underline{16.8} & \underline{28.1} & 4.0  & \underline{18.9} \\
UI-TARS-2B~\cite{qin2025ui} & $47.4$ & $4.1$ & $26.4$ & $42.9$ & $6.3$ & $27.6$ & $17.8$ & $4.7$ & $14.6$ & $56.9$ & $17.3$ & $39.8$ & $50.3$ & $17.0$ & $42.6$ & $21.5$ & $5.6$ & $14.3$ & $39.6$ & $8.4$ & $27.7$ \\
UI-TARS-7B~\cite{qin2025ui}    & $58.4$ & $12.4$ & $36.1$ & $50.0$ & $9.1$ & $32.8$ & $\mathbf{20.8}$ & $9.4$ & $\mathbf{18.0}$ & $63.9$ & $\mathbf{31.8}$ & $\mathbf{50.0}$ & $\mathbf{63.3}$ & $20.8$ & $53.5$ & $30.8$ & $\mathbf{16.9}$ & $24.5$ & $47.8$ & $16.2$ & $35.7$ \\
\hline
Qwen2.5-VL-3B $\dagger$ & 14.9 & \underline{2.1} & 8.7 & 7.1 & 2.1 & 5.0 & 1.5 & 1.6 & 1.5 & 13.9 & \underline{8.2} & 11.4 & 10.7 & 3.8 & 9.1 & 9.4 & 2.3 & 6.1 & 9.1 & 3.3 & 6.9\\
Qwen2.5-VL-7B $\dagger$ & 22.1 & 0.7 & 11.7 & 16.2 & \underline{4.2} & 11.1 & 5.1 & 3.1 & 4.6 & 25.0 & 7.3 & 17.3 & 22.0 & \underline{11.3} & 19.6 & 17.8 & \underline{4.5} & 11.7 & 17.4 & \underline{4.5} & 12.5\\
TongUI-3B & 32.5 & 0.7 & 17.1 & 24.8 & 2.8 & 15.5 & 11.7 & 1.6 & 9.2 & \underline{43.1} & \textbf{12.7} & \underline{29.9} & 32.8 & 7.6 & 27.0 & 15.0 & 1.1 & 8.7 & 26.4 & 4.1 & 18.0\\
TongUI-7B & \textbf{40.9} & \textbf{3.5} & \textbf{22.7} & \textbf{31.3} & \textbf{7.0} & \textbf{21.1} & \textbf{17.3} & \textbf{9.4} & \textbf{15.3} & \textbf{50.7} & \textbf{12.7} & \textbf{34.3} & \textbf{45.8} & \textbf{13.2} & \textbf{38.3} & \textbf{28.0} & \textbf{6.7} & \textbf{18.4} & \textbf{35.1} & \textbf{8.0} & \textbf{24.7}\\
TongUI-32B & 44.8 & 6.2 & 26.1 & 45.5 & 7.7 & 29.6 & 26.4 & 6.3 & 21.5 & 61.8 & 27.3 & 46.9 & 61.8 & 27.3 & 48.3 & 45.8 & 11.2 & 30.1 & 45.9 & 12.6 & 33.1\\
\hline
\end{tabular}%
}
\caption{Results on ScreenSpot-Pro.}
\label{table:supp_experiment_ScreenSpotpro}
\end{table*}

\section{More Experiments}

\begin{table}
\begin{center}
\small
\begin{tabular}{c | c | c c c}
\hline
\multirow{2}{*}{\bf Data} & \multirow{2}{*}{\emph{ScreenSpot}} & \multicolumn{3}{c}{\emph{Mind2Web}} \\
& & \multicolumn{1}{c}{\emph{Task}}&\multicolumn{1}{c}{\emph{Website}}&\multicolumn{1}{c}{\emph{Domain}}\\
\hline 
No SFT & 56.5 & 0.4 & 1.0 & 1.7 \\
Refined data & 68.0 & 39.7 & 35.5 & 40.7 \\
+ WikiHow 50K & 75.8 & 42.1 & 39.6 & 44.4 \\
+ Baidu 50K & 78.7 & 43.4 & 41.6 &  45.5 \\
+ Video 50K & {79.6} & {44.2} & {42.6} &  {46.0} \\
+All data & \textbf{83.6} & \textbf{48.8} & \textbf{48.1} & \textbf{49.5}\\
\hline
\end{tabular}
\end{center}
\caption{Performance scaling over data sources, where \emph{Step SR} is reported for Mind2Web.}
\label{table:experiment_Scaling}
\end{table}

\subsection{Grounding Details}

We provide detailed results on ScreenSpot, ScreenSpot-V2, and ScreenSpot-Pro, as shown in~\cref{table:supp_experiment_ScreenSpot,table:supp_experiment_ScreenSpotV2,table:supp_experiment_ScreenSpotpro}, respectively.

\begin{table}
\begin{center}
\small
\begin{tabular}{c|c|c c c|c}
\hline
\textbf{Setting} & \textbf{Metric} & Run 1 & Run 2 & Run 3 & Mean ± Std \\
\hline
\multirow{3}{*}{Cross-Task} 
    & \emph{Elem. Acc} & 52.0 & 52.9 & 52.5 & 52.5 ± 0.451 \\
    & \emph{OP. F1}    & 88.6 & 88.6 & 88.8 & 88.7 ± 0.115 \\
    & \emph{Step SR}  & 47.6 & 48.5 & 48.0 & 48.0 ± 0.451 \\
\hline
\multirow{3}{*}{Cross-Web} 
    & \emph{Elem. Acc} & 52.7 & 51.8 & 52.4 & 52.3 ± 0.458 \\
    & \emph{OP. F1}    & 86.0 & 86.3 & 86.6 & 86.3 ± 0.300 \\
    & \emph{Step SR}  & 46.1 & 45.7 & 46.5 & 46.1 ± 0.400 \\
\hline
\multirow{3}{*}{Cross-Domain} 
    & \emph{Elem. Acc} & 54.1 & 53.8 & 53.6 & 53.8 ± 0.252 \\
    & \emph{OP. F1}    & 88.7 & 88.5 & 88.4 & 88.5 ± 0.153 \\
    & \emph{Step SR}  & 49.7 & 49.6 & 49.4 & 49.6 ± 0.153 \\
\hline
\end{tabular}
\end{center}
\caption{Error bar on Mind2Web.}
\label{tab:error-bar}
\end{table}

\begin{table}[H]
\centering
\small
\begin{tabular}{c c | c c }
\toprule
\textbf{Method} & \textbf{Model} & \emph{High} & \emph{Low} \\
\midrule
LT-5$\dagger$& PaLM 2S & 30.3 & 57.1 \\
LT-10$\dagger$ & PaLM 2S & 28.5 & 58.9 \\
LT-100$\dagger$ & PaLM 2S & 39.8 & 62.8 \\
LT-1k$\dagger$ & PaLM 2S & 52.5 & 71.4 \\
LT-10k$\dagger$ & PaLM 2S & 62.0 & 84.7 \\
LT-all$\dagger$ & PaLM 2S & 65.6 & 81.8 \\
LT-1k-r64$\dagger$ & PaLM 2S & 54.8 & 76.6 \\
LT-10k-r64$\dagger$ & PaLM 2S & 69.6 & 81.8 \\
LT-all-r64$\dagger$ & PaLM 2S & 71.5 & 86.6 \\
\hline
TongUI & TongUI-3B & \underline{73.3} & \underline{91.5} \\
TongUI & TongUI-7B & \textbf{76.0} & \textbf{91.9} \\
\bottomrule
\end{tabular}
\vspace{5pt}
\caption{Step accuracy on AndroidControl. Experiments marked with $\dagger$ is from ~\cite{li2025effects}. ``LT-10k-r64'' means the model is tuned by LoRA with 10k training trajectories and rank 64.}
\label{table:experiment_AndroidControl_tuned}
\end{table}

\begin{table}[H]
\centering
\small
\begin{tabular}{c c | c }
\toprule
\textbf{Model} & \textbf{FT?} & \emph{Step Accuracy} \\
\midrule
CogAgent~\cite{hong2024cogagent} & No & 11.8\\
GPT-4V$\dagger$ & No & 18.76 \\
GPT-4O$\dagger$ & No & 20.39 \\
Qwen-VL-Chat(7B)$\dagger$ & Yes & 72.81\\
OdysseyAgent$\dagger$ & Yes & \underline{74.25} \\
\hline
TongUI-3B & Yes & 71.04\\
TongUI-7B & Yes & \textbf{74.81} \\
\bottomrule
\end{tabular}
\vspace{5pt}
\caption{Step accuracy on GUI Odyssey. ``FT?'' indicates if the model is fine-tuned on GUI Odyssey training set. Experiments mared with $\dagger$ is reported by previous work \cite{lu2024gui}}
\label{table:experiment_GUIOdyssey_tuned}
\end{table}

\begin{table}[t]
\begin{center}
\small
\begin{tabular}{c | c c}
\hline
\multicolumn{1}{c}{\bf Model} & \multicolumn{1}{c}{\emph{Action Accuracy}} & \multicolumn{1}{c}{\emph{Success Rate}} \\ \hline 
ShowUI-2B\cite{lin2024showui} & 65.7 & 14.7 \\
Qwen2.5-VL-3B & 36.3 & 10.8\\
Qwen2.5-VL-7B & 70.6 & 16.6 \\
TongUI-3B & 86.3 & 18.6 \\
TongUI-7B & \underline{87.3} & \underline{25.5} \\
TongUI-32B & \textbf{88.2} & \textbf{56.9} \\
\hline
\end{tabular}
\end{center}
\caption{Results on the Baidu Experience}
\label{table:experiment_baidu}
\end{table}

\subsection{Data Scaling Effect}

We evaluate the scaling law of GUI agents using collected data from different sources with Qwen2.5-VL-7B. 
Concretely, (1) we report the performance of w/o SFT; (2) we use the refined public data to tune the model; (2) we add data collected from WikiHow to tune the model; (4) we add data collected from Baidu Experience to tune the model; (5) we add data collected from Video tutorials to tune the model. Results is shown in~\cref{table:experiment_Scaling}. We find that with the increase of data numbers and data sources, the performance of GUI agents is also constantly improving.

\subsection{Error Bar}

To evaluate the robustness and stability of our model, we conduct three independent runs on Mind2Web for each type of tasks: Cross-Task, Cross-Web, and Cross-Domain, using TongUI-3B. Table~\ref{tab:error-bar} reports the performance on three evaluation metrics(Element Accuracy, Operation F1, and Step Success Rate) along with their mean and standard deviation (Mean ± Std) across runs. The results show that the model exhibits consistent performance with low variance, indicating stable performance of our model.

\subsection{Results on AndroidControl}
In \cref{table:experiment_AndroidControl_tuned}, we show the model performance of our model fine-tuned in the AndroidControl training set. The result indicates that our model achieves the best performance on both low and high level instructions compared to other models trained on the same training set.


\subsection{Results on GUI Odyssey }
In \cref{table:experiment_GUIOdyssey_tuned}, we show the results evaluated on GUI Odyssey Test-Random split. TongUI-7B achieves the best performance compared with open-source and closed-source models.

\section{Examples}

\subsection{GUI-Net}

\begin{figure*}
    \centering
    \includegraphics[width=0.98\linewidth]{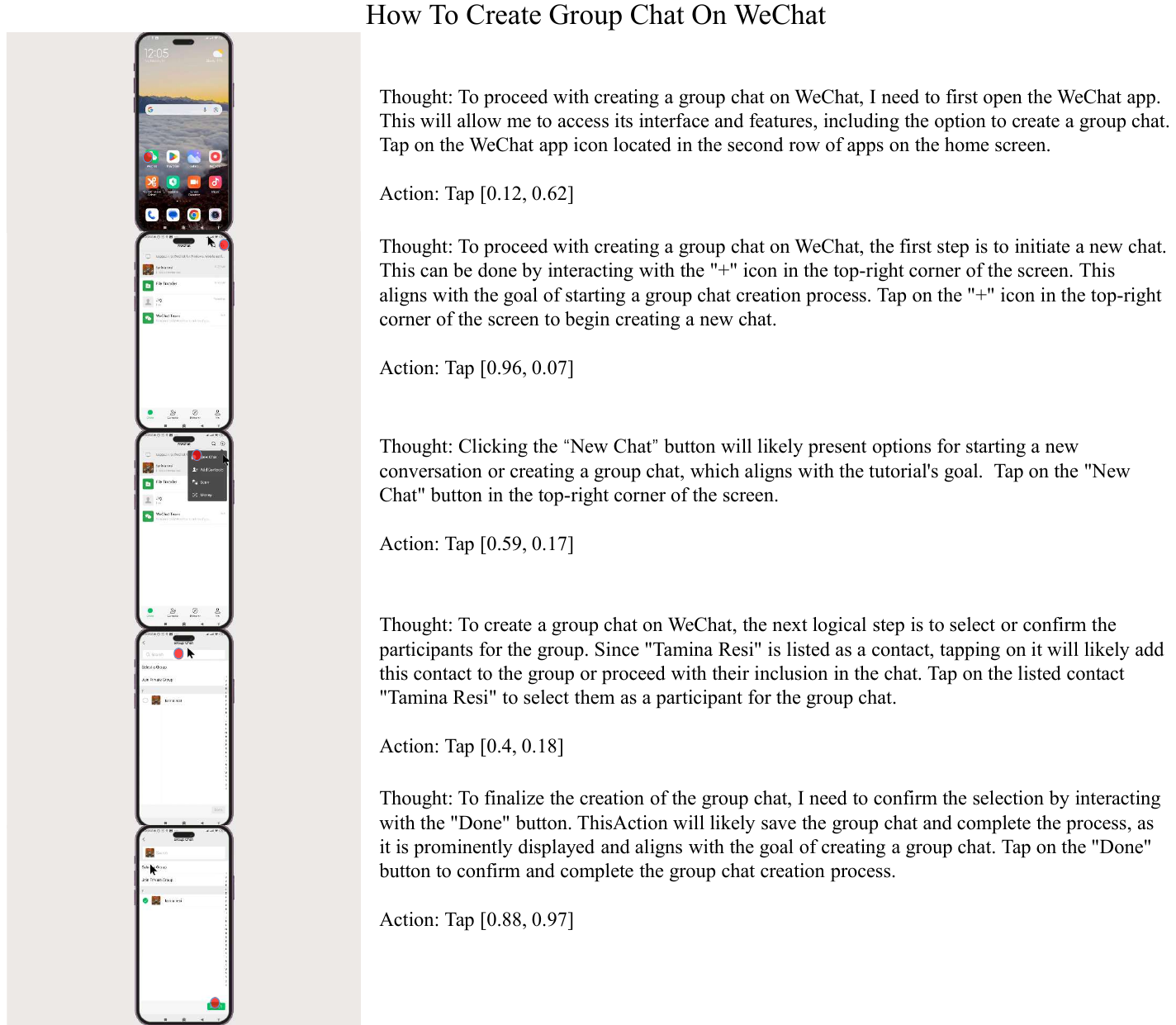}
    \caption{Dataset example.}
    \label{fig:dataset1}
\end{figure*}

\begin{figure*}
    \centering
    \includegraphics[width=0.98\linewidth]{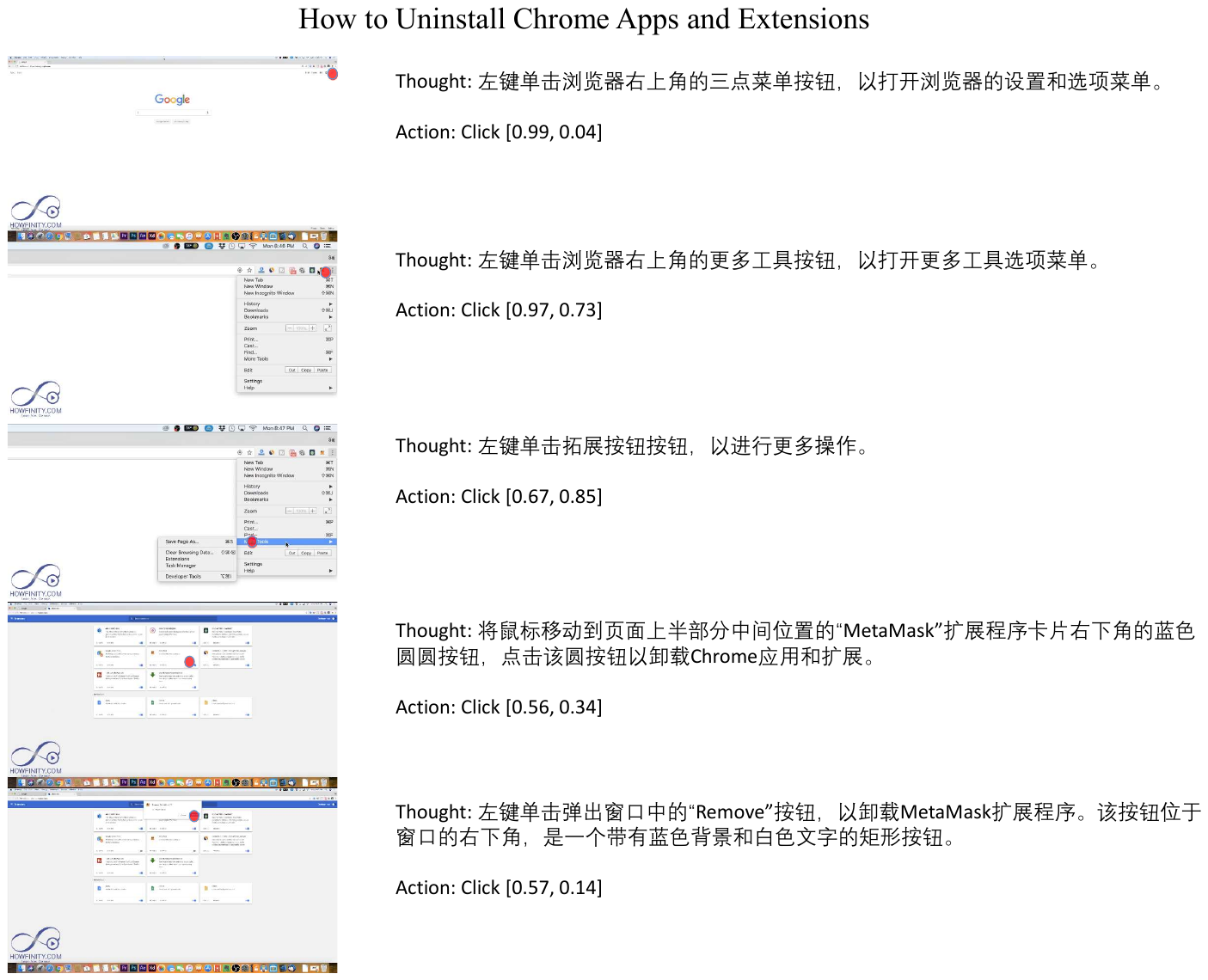}
    \caption{Dataset example.}
    \label{fig:dataset2}
\end{figure*}

\begin{figure*}
    \centering
    \includegraphics[width=0.98\linewidth]{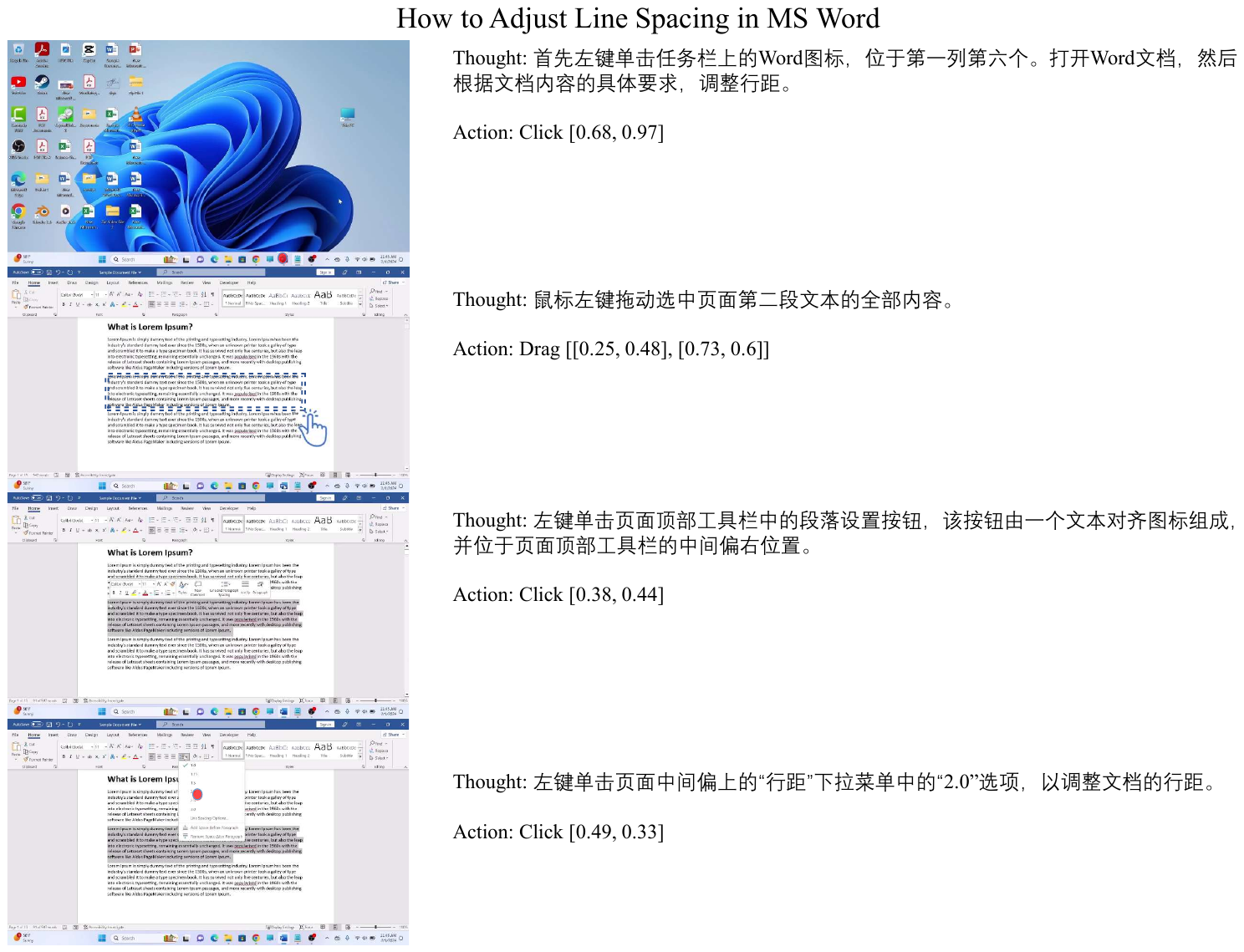}
    \caption{Dataset example.}
    \label{fig:dataset3}
\end{figure*}

\begin{figure*}
    \centering
    \includegraphics[width=0.98\linewidth]{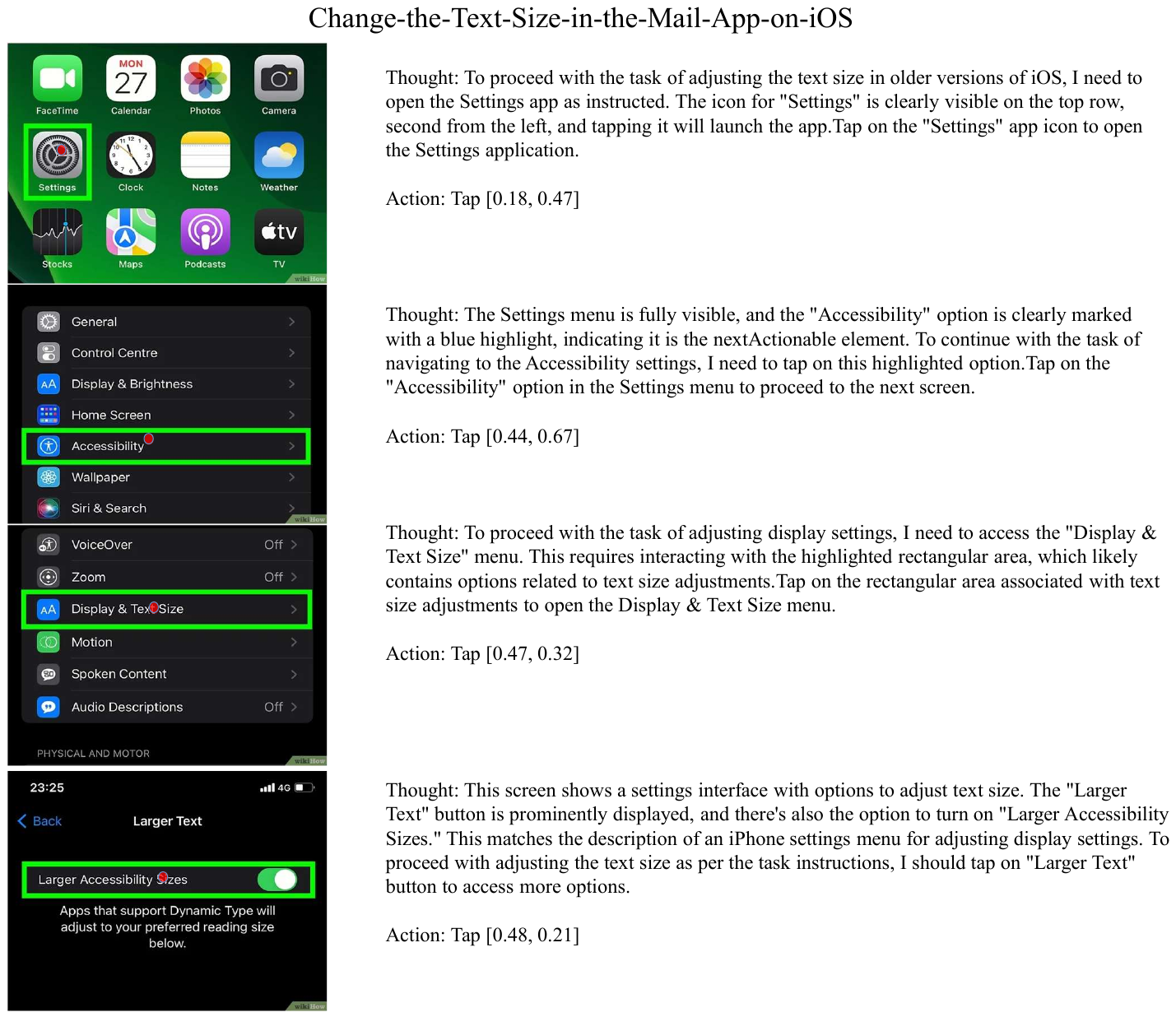}
    \caption{Dataset example.}
    \label{fig:dataset4}
\end{figure*}

\begin{figure*}
    \centering
    \includegraphics[width=0.98\linewidth]{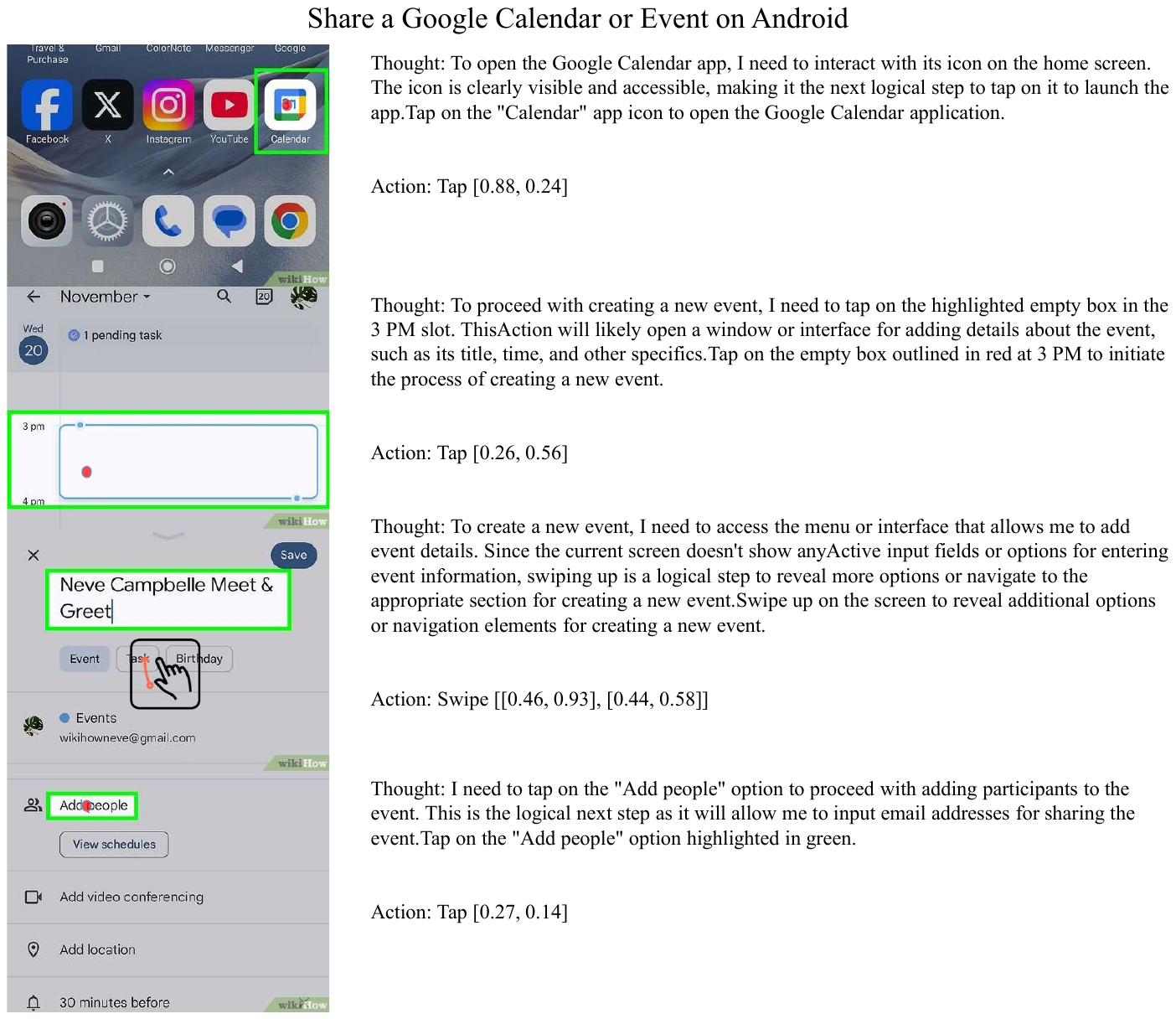}
    \caption{Dataset example.}
    \label{fig:dataset5}
\end{figure*}

We visualize several examples of collected trajectory data, as shown in~\cref{fig:dataset1,fig:dataset2,fig:dataset3,fig:dataset4,fig:dataset5}.

\subsection{TongUI}

\begin{figure*}
    \centering
    \includegraphics[width=0.98\linewidth]{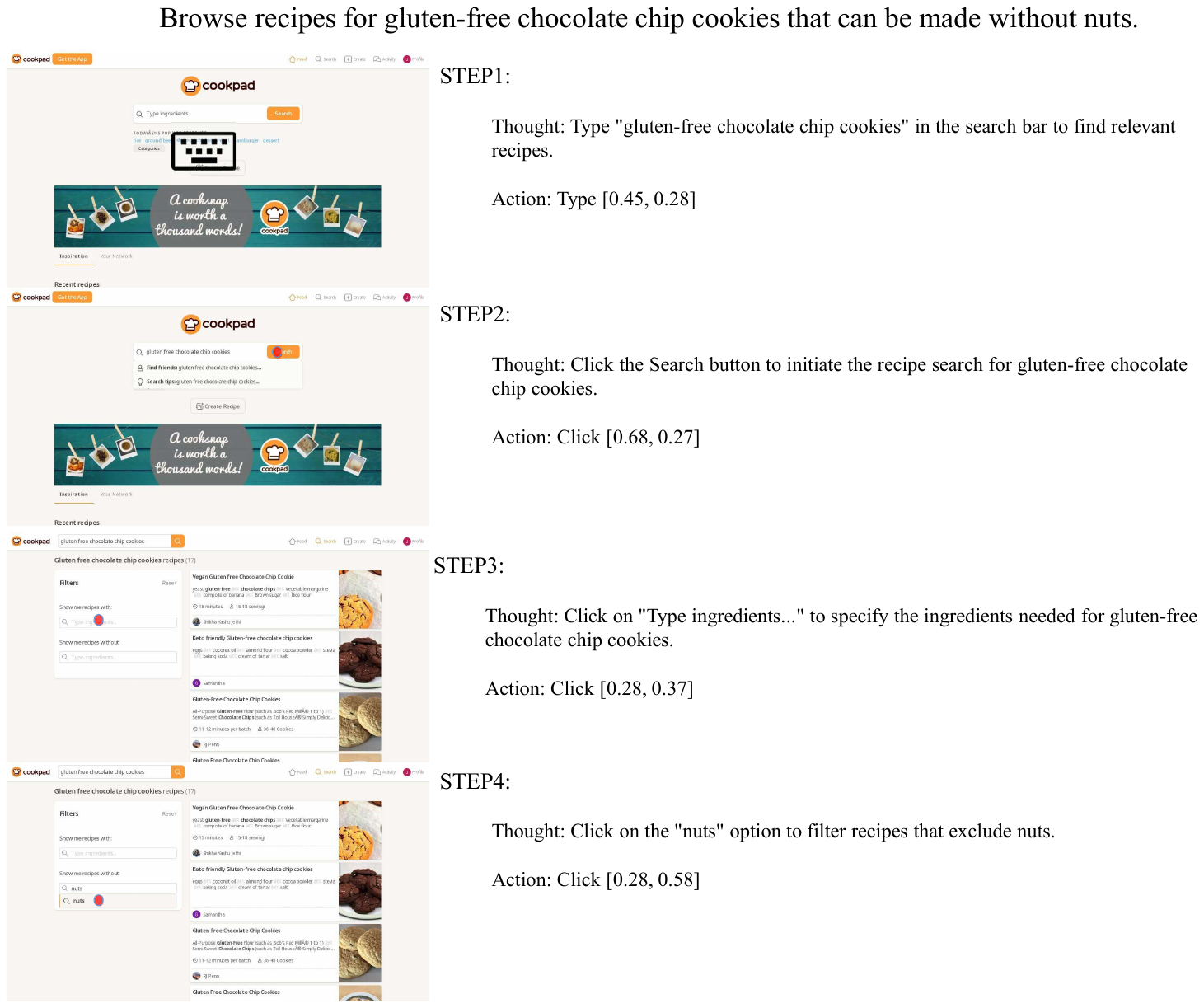}
    \caption{Examples executed by TongUI.}
    \label{fig:model1}
\end{figure*}

\begin{figure*}
    \centering
    \includegraphics[width=0.98\linewidth]{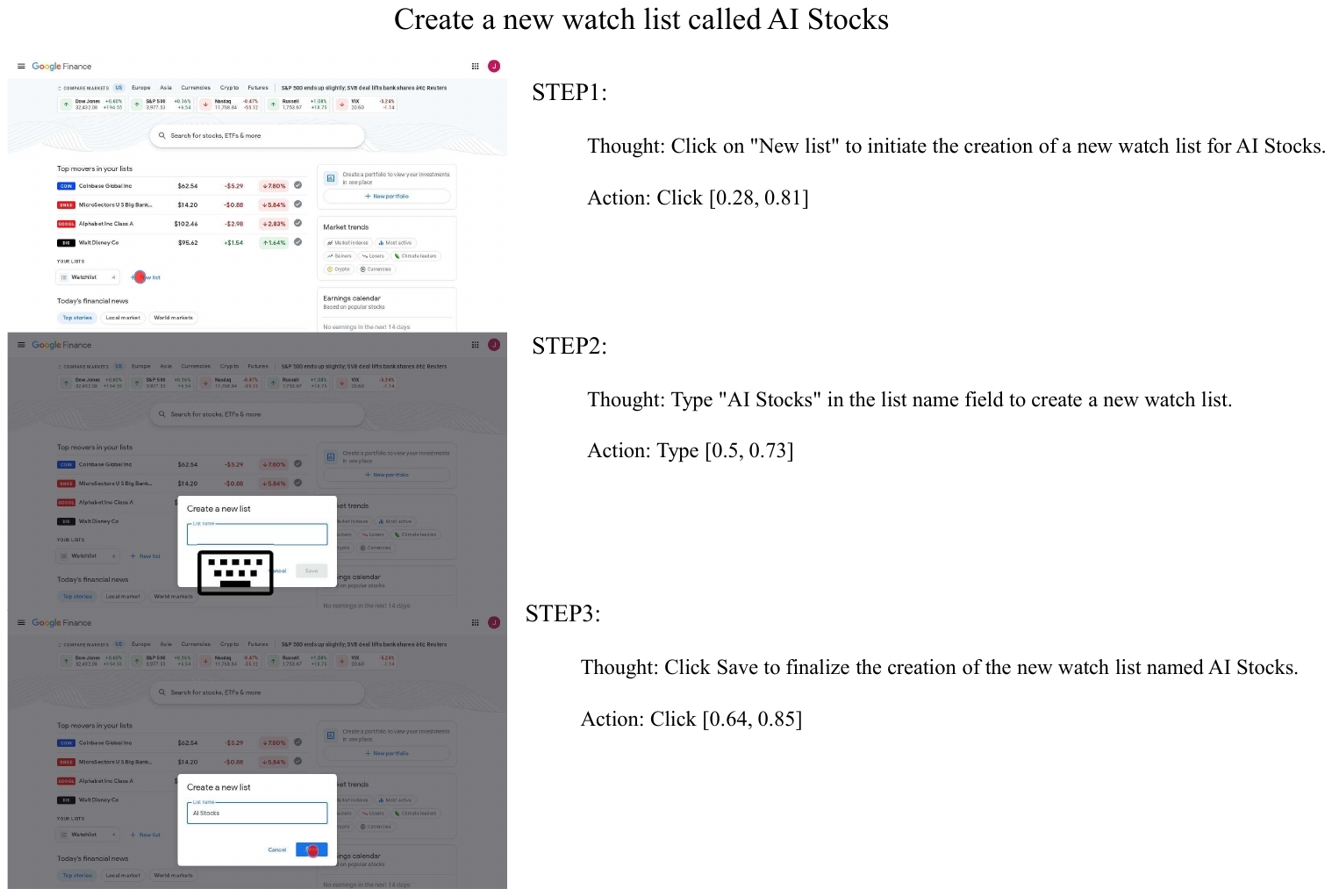}
    \caption{Examples executed by TongUI.}
    \label{fig:model2}
\end{figure*}

\begin{figure*}
    \centering
    \includegraphics[width=0.98\linewidth]{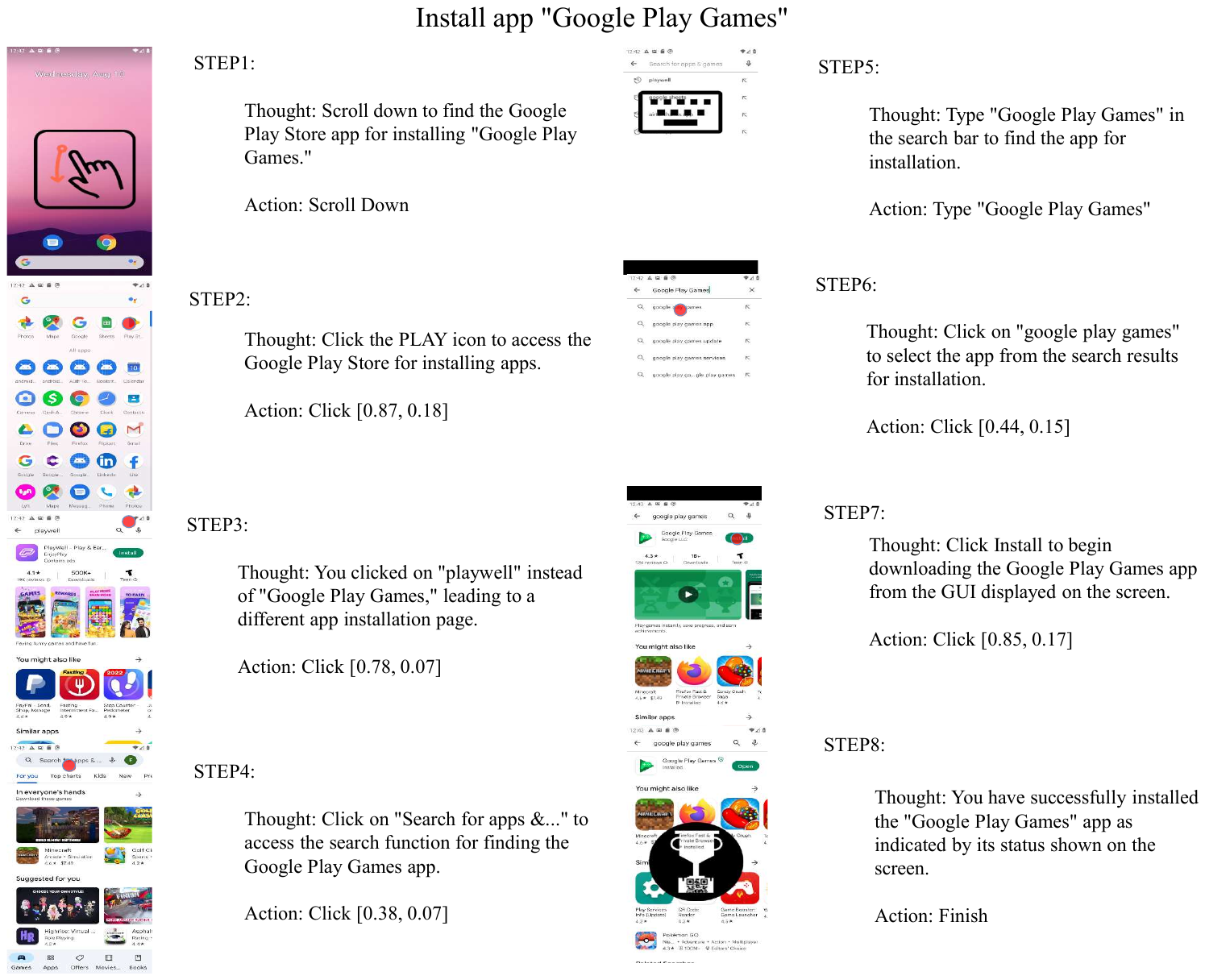}
    \caption{Examples executed by TongUI.}
    \label{fig:model3}
\end{figure*}

\begin{figure*}
    \centering
    \includegraphics[width=0.98\linewidth]{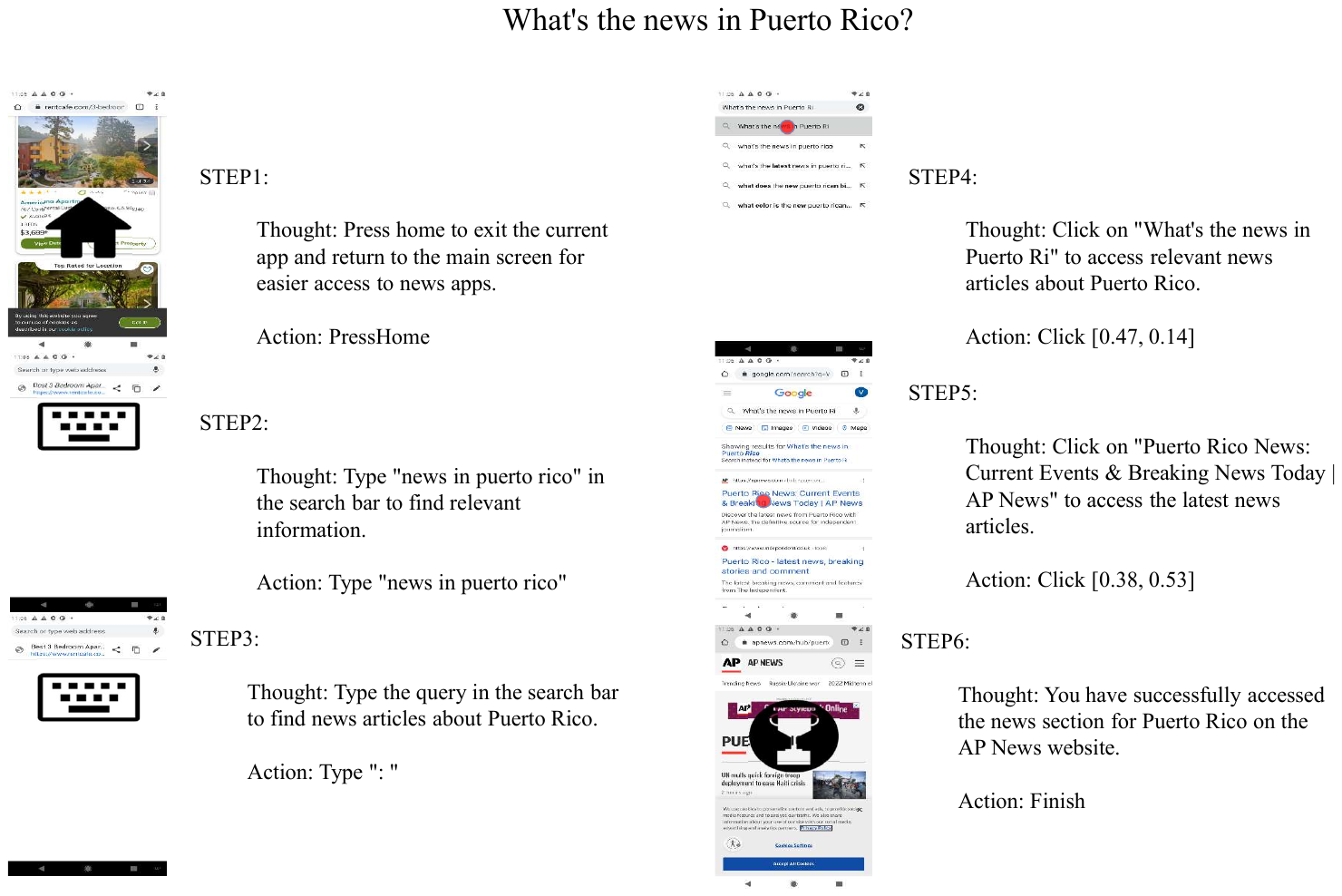}
    \caption{Examples executed by TongUI.}
    \label{fig:model4}
\end{figure*}

We visualize several examples executed by the TongUI agent, as shown in~\cref{fig:model1,fig:model2,fig:model3,fig:model4}.

